\documentclass[10pt,journal,compsoc]{IEEEtran}
\ifCLASSOPTIONcompsoc
  \usepackage[nocompress]{cite}
\else
  \usepackage{cite}
\fi

\usepackage{microtype}
\usepackage[T1]{fontenc}
\usepackage[utf8]{inputenc}
\usepackage[french,main=english]{babel}
\usepackage[pagebackref,breaklinks,colorlinks]{hyperref}
\usepackage{amsmath,amssymb,amsthm,amsfonts}
\usepackage{booktabs}  
\usepackage{multirow}
\usepackage{graphicx}
\usepackage{float}
\usepackage{graphbox}
\usepackage{caption}
\usepackage{subcaption}
\usepackage[ruled,vlined,linesnumbered]{algorithm2e}
\usepackage[export]{adjustbox}
\usepackage{threeparttable}
\usepackage{tabularx}
\usepackage{url} 
\usepackage{csquotes}

\newcommand\modelname{Document Attention Network}
\newcommand\modelacc{DAN}

\usepackage{pifont}

\newcommand{\xmark}{\ding{55}}

\newcommand{\mb}[1]{\boldsymbol{#1}}
\newcommand{\mset}[1]{\mathcal{#1}}
\newcommand{\mseq}[1]{\mb{#1}}

\title{\modelacc{}: a Segmentation-free \modelname{} for Handwritten Document Recognition}

\begin{document}
\sloppy

\author{Denis~Coquenet,
        Clément~Chatelain,
        and~Thierry~Paquet
\IEEEcompsocitemizethanks{
\IEEEcompsocthanksitem D. Coquenet and T. Paquet are with LITIS EA 4108, University of Rouen Normandy and Normandie Université, France.\\
E-mail: \{denis.coquenet,thierry.paquet\}@litislab.eu
\IEEEcompsocthanksitem C. Chatelain is with LITIS EA 4108, INSA Rouen Normandy and Normandie Université, France.\\
E-mail: clement.chatelain@litislab.eu
}%
\thanks{Manuscript received ...; revised ...}}

\IEEEtitleabstractindextext{
\begin{abstract}
Unconstrained handwritten text recognition is a challenging computer vision task. It is traditionally handled by a two-step approach, combining line segmentation followed by text line recognition. For the first time, we propose an end-to-end segmentation-free architecture for the task of handwritten document recognition: the \modelname{}. In addition to text recognition, the model is trained to label text parts using begin and end tags in an XML-like fashion. This model is made up of an FCN encoder for feature extraction and a stack of transformer decoder layers for a recurrent token-by-token prediction process. It takes whole text documents as input and sequentially outputs characters, as well as logical layout tokens. Contrary to the existing segmentation-based approaches, the model is trained without using any segmentation label. We achieve competitive results on the READ 2016 dataset at page level, as well as double-page level with a CER of 3.43\% and 3.70\%, respectively. We also provide results for the RIMES 2009 dataset at page level, reaching 4.54\% of CER.

We provide all source code and pre-trained model weights at \url{https://github.com/FactoDeepLearning/DAN}.

\end{abstract}

}
\IEEEdisplaynontitleabstractindextext
\IEEEpeerreviewmaketitle
\maketitle

\section{Introduction}
\IEEEPARstart{A}{} handwritten document is a complex structure composed of handwritten text blocks structured according to a specific layout. Handwritten Document Recognition (HDR) can be defined as the joint recognition of both text and layout. To our knowledge, this work is the first attempt to build an end-to-end approach for HDR. Until now, the previous works have only focused on Handwritten Text Recognition (HTR) of isolated text blocks, or Document Layout Analysis (DLA), considering them as two independent tasks. 

Offline HTR consists in recognizing the text from a digitized document. HTR at document level has always been relying on a prior segmentation stage to extract the text blocks from the input image (at character, word, then line or even paragraph level), before recognizing the text itself. Such two-step approach (segmentation + recognition) has several drawbacks. It relies on segmented entities, which do not have a clear definition from the image point of view. For instance, text lines can be defined by X-heights, \textit{i.e.}, as the area that covers the core of the text, without its ascenders and descenders; but, it can also be defined by baselines, bounding boxes, or polygons \cite{Renton2018}. This approach also requires segmentation annotations, which are very costly to produce.
The resulting predictions accumulate errors from both stages. Another point is about the reading order. Since the segmentation stage is generally a one-shot process, there is no notion of ordered sequence between the different text regions, which prevents learning the reading order. 

\begin{figure}[t!]
    \centering
    \begin{subfigure}[b]{\linewidth}
    \includegraphics[width=\linewidth]{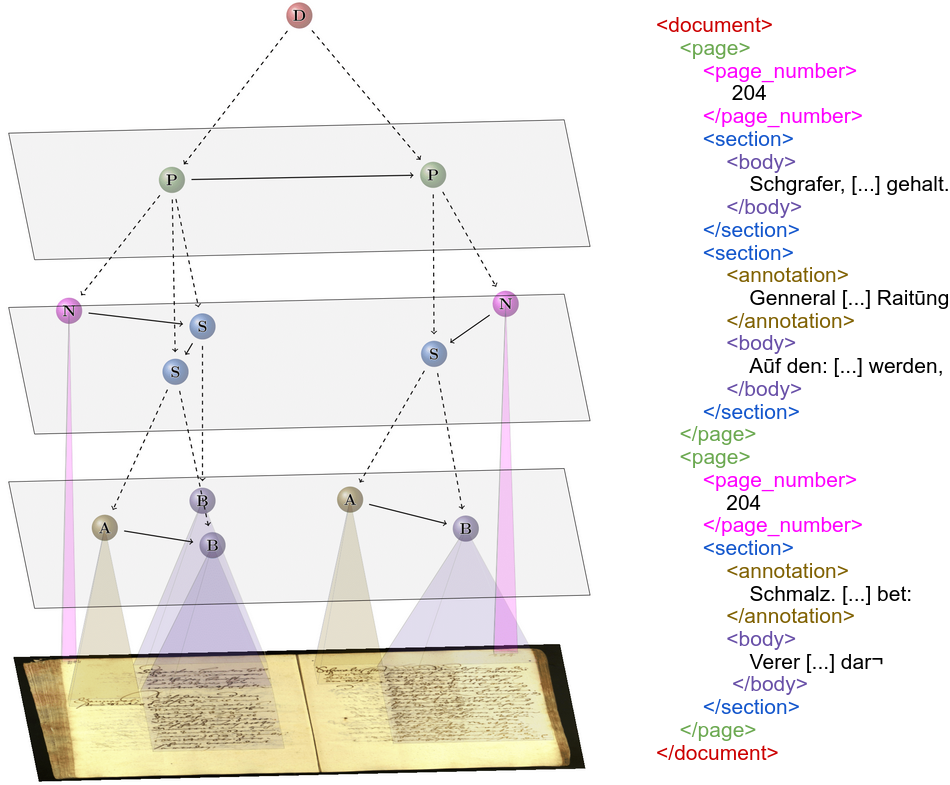}
    \caption{Left: document image with associated layout graph. The layout entities are represented by the nodes. Dashed arrows correspond to a membership relation, while solid arrows indicate the reading order. Right: the ground truth is a serialized representation of the document which follows the XML paradigm, including both text and layout tokens.}
    \label{fig:xml}
    \end{subfigure}
    \par\bigskip
    \begin{subfigure}[b]{\linewidth}
    \includegraphics[width=\linewidth,frame]{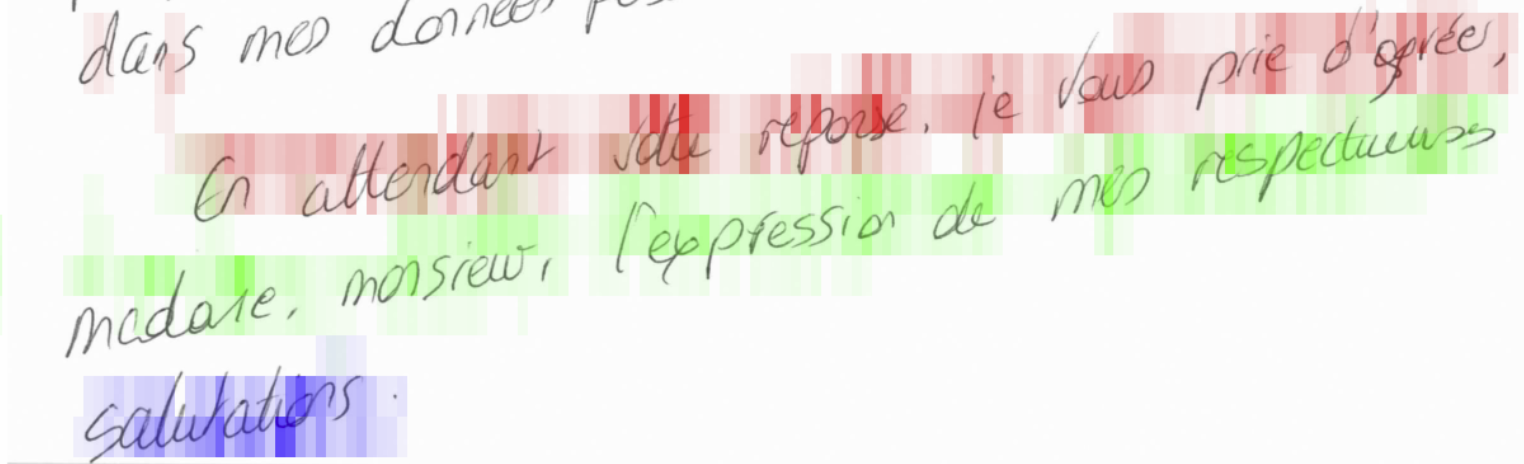}
    \noindent \scriptsize{Prediction: "En attendant votre réponse, je vous prie d'agréer, Madame, Monsieur, l'expression de mes respectueuses salutations."}
    \caption{Attention weights visualization for a validation sample of the RIMES 2009 dataset. The character-level attention enables to follow the slant of the text lines.}
    \label{fig:viz_slanted_lines}
    \end{subfigure}
    \caption{Visualization of input, ground truth and prediction.}
\end{figure}

In the past, the emergence of end-to-end models enabled to alleviate the need for segmentation labels. The recently proposed approaches \cite{Bluche2016,yousef2020,span,van} show that it is possible to recognize text from paragraph images without any explicit text line segmentation step. However, these models are limited to the recognition of single columns of text. It means that they can only be used for simple layouts or used as a second step, after a prior paragraph segmentation step. However, real-world documents have more complex layouts most of the time, including multiple columns of text or annotations in the margin. In this paper, we propose to process whole documents, with such complex layouts, recognizing both text and layout. The proposed approach consists of a unified statistical model which only needs very few annotations for training, without needing any segmentation (or physical) label.

One example of such a complex (real) document is depicted in Figure \ref{fig:xml}. While the textual content can be represented as a sequence of characters, the layout is represented as an oriented graph, in order to model the hierarchy and the reading order of the different layout entities. In figure \ref{fig:xml}, this layout graph is projected on the document image: nodes are layout entities and edges model the relations between them. A membership relation is represented by a dashed arrow, while solid arrows represent the reading order. In this example, the document is made up of two pages, each page containing a page number and a sequence of sections, each section being made up of zero, one or many marginal annotations and a single body. 

Text and layout are intrinsically linked: text recognition may help to label a layout entity, and vice versa. Therefore, we have turned toward the joint recognition of text and layout in a unique model. As shown on the right side of Figure \ref{fig:xml}, we chose the XML paradigm to generate a serialized representation of the document that is further used as the ground truth of this document. Notice that no physical information is encoded in the ground truth. Contrary to the standard document image analysis tasks, the proposed approach does not rely on segmentation labels of text regions such as bounding boxes, for instance. Instead, the proposed model is able to provide transcriptions enriched with logical layout information, leading to structured transcriptions. It notably enables to reduce the need for costly segmentation annotations.

The proposed model recurrently predicts the character and layout tokens through a character-level attention mechanism. It enables to deal with slanted lines, as shown in Figure \ref{fig:viz_slanted_lines}.
Processing whole documents instead of isolated paragraphs raises new challenges:
\begin{itemize}
    \item Paragraph-level models usually benefit from the fact that all horizontal elements at the same vertical position belong to the same text line. This assumption is not always valid at document-level.
    \item The input images are larger and the associated target sequences are longer for documents than for paragraphs, leading to more complex training procedures, especially in the case of attention-based models, which involve a growing need for GPU memory. 
    \item While paragraphs are read in a monotonous order (characters are read from left to right and lines from top to bottom in case of common occidental languages for example), documents reading order is layout dependent, \textit{i.e.}, paragraphs of a single-column document are read from top to bottom only, whereas a multi-column document is commonly read column by column, adding a horizontal constraint on the reading order.
\end{itemize}

The transformer architecture \cite{Vaswani2017} was first introduced in the context of machine translation. Since then, transformer-based models have proven their robustness through a wide variety of computer vision tasks, such as images of mathematical expression recognition \cite{zhao2021} or image classification \cite{Dosovitskiy2021}.
They have also shown promising results for single text line recognition \cite{wick2021,kang2020} and scene text recognition \cite{Atienza2021}. 

In this work, we propose a \modelname{} (\modelacc{}) based on an FCN encoder and a transformer decoder for handwritten document recognition. 

In brief, we make the following contributions:
\begin{itemize}
    \item We propose the \modelname{}: the first end-to-end architecture able to recognize text at document level while labeling logical layout information, in a single decoding process. 
    
    \item The model is trained in an end-to-end fashion without the need for any segmentation labels.
    
    \item We provide the first results on the RIMES 2009 dataset at page and paragraph levels, as well as on the READ 2016 dataset at page and double-page levels. These READ 2016 results are competitive compared to state-of-the-art paragraph-level and line-level approaches, while the proposed approach does not need a prior segmentation step.
    
    \item The proposed network achieves state-of-the-art results on the RIMES 2011 dataset at paragraph level and on the READ 2016 dataset at line and paragraph levels.
    
    \item We propose two new metrics to evaluate the quality of the recognized document layout.
    
    \item We provide all source code, including metrics and synthetic document generation process, as well as the trained model weights.
    
\end{itemize}

This paper is organized as follows. Section \ref{section-related-works} is dedicated to the related works. The architecture and the training strategy are presented in Section \ref{section-architecture}. We present the proposed metrics in Section \ref{section-metrics}. Experiments are detailed in Section \ref{section-experiments}, and the limitations are presented in Section \ref{section-limitations}. We provide a discussion in Section  \ref{section-discussion}, and we draw a general conclusion in Section \ref{section-conclusion}.
\section{Related Works}
\label{section-related-works}

Deep learning models are becoming more and more powerful and can now process entire documents. As we aim at recognizing both the text and the logical layout information of a handwritten document image, the task falls into both the handwriting recognition field and the layout analysis field, thus sharing links with document understanding. Therefore, this section is first dedicated to document understanding in a general way and then focuses on layout analysis and handwriting recognition. 

\subsection{Document understanding}
Document understanding includes a set of tasks whose purpose is to extract, classify, interpret, contextualize, and search information from documents. It implies, among others, document layout analysis and Optical Character Recognition (OCR). But it also consists in understanding complex structures such as tables, schemes, or images. This is a developing field and here are some related works.
Katti \textit{et al.} \cite{Chargrid} and Reisswig \textit{et al.} \cite{Chargrid_OCR} proposed Chargrid, a way to represent textual documents as a 2D representation of one-hot encoded characters, which are produced by an OCR. The idea is to keep the spatial information between the textual entities. Chargrid is then used as the input for a Key Information Extraction (KIE) task, implying three sub-tasks: bounding box regression, semantic segmentation and box masking. 
In \cite{Bertgrid}, the Chargrid paradigm is also used, but the character's encoding is superseded by word embeddings through the use of the BERT model \cite{BERT}.
Xu \textit{et al.} \cite{LayoutLMv2} tackle the task of Visually-rich Document Understanding (VDU). They used a transformer architecture applied to multiple modalities: OCR text and bounding boxes, and visual embedding. Powalski \textit{et al.} \cite{Tilt} proposed a transformer-based model able to tackle multiple tasks such as document classification, KIE and Visual Question-Answering (VQA).
All these works imply the use of large datasets, mainly synthetic: they need a lot of information either as input or as ground truth annotation (notably for bounding boxes).
One can notice that document understanding is still in its early stages. It mainly focuses on 2D document representation and information extraction. Our work differs from document understanding in that we propose to extract all the text from a document; we do not aim at retrieving only specific information. 

\subsection{Document layout analysis}
Document Layout Analysis (DLA) aims at identifying and categorizing the regions of interest in a document image. Barlas \textit{et al.} \cite{Barlas2014} proposed a method based on connected components to recognize 8 types of objects on heterogeneous documents: text, photographic image, hand-drawn line area, graph, table, edge line, separator and material damage. Quirós \textit{et al.} \cite{Quiros2018} proposed a 2-stage method based on an artificial neural network for the task of semantic segmentation on historical handwritten documents. It detects multiple zone types such as page numbers, marginal notes and main paragraphs. 

Currently, there are mainly two approaches to handle DLA: pixel-by-pixel classification and bounding box prediction. 

Fully Convolutional Networks (FCN) are the most popular approach for pixel-level DLA (\cite{Soullard2020,DHSegment,Yang2017}). It consists of an elegant and relatively light end-to-end model that does not require re-scaling the input images. Yang \textit{et al.} \cite{Yang2017} applied DLA on printed textual documents: contemporary magazines and academic papers. They trained their model to recognize multiple classes namely figures, tables, section headings, captions, lists, and paragraphs. The model presented in \cite{DHSegment} aims at detecting different items from historical documents: text regions, decorations, comments, and background. In \cite{Soullard2020}, the model is applied to historical newspapers. It recognizes many textual elements such as titles, text blocks and advertisements, as well as images.
Renton \textit{et al.} \cite{Renton2018}, Oliveira \textit{et al.} \cite{DHSegment}, Grüning \textit{et al.} \cite{ARUNet}, and Boillet \textit{et al.} \cite{Boillet2020,Boillet2022} focused on text line segmentation, as a first step before text line recognition, also using FCN.

Object-detection approaches for word bounding box predictions have been studied in \cite{Carbonell2020,Chung2020}. They follow the standard object-detection paradigm based on a region proposal network and a non-maximum suppression algorithm \cite{Liu2016}.
Tensmeyer \textit{et al.} \cite{Wigington2019} and Moysset \textit{et al.} \cite{Moysset2017} focused on start-of-line prediction to extract normalized lines. 

DLA focuses on identifying physical regions of interest, whether they are textual or not. It is driven by physical ground truth annotations accounting for semantic labels that are associated to document logical elements. In this paper, we focus on the textual components only, for which we target their recognition and semantic labeling.

\subsection{Handwritten text recognition}
Most of the works on handwriting recognition deal with images of isolated lines or words. 
It implies the use of a prior line segmentation step, carried out by a DLA stage.

For the text line images recognition task itself, many architectures have been proposed: Multi-Dimensional Long Short Term Memory (MD-LSTM) \cite{Voigtlaender2016}, combination of Convolutional Neural Networks (CNN) and LSTM \cite{Wigington2017} (also used for scene text recognition \cite{Shi2017}), CNN \cite{Coquenet2019} and FCN \cite{Coquenet2020}. During training, they all rely on the CTC \cite{CTC} loss to handle the sequence alignment issue induced by the variability of the input image widths and the target sequence lengths. In \cite{Michael2019}, the CTC loss is superseded by the cross-entropy loss, and the sequence alignment issue is handled by an attention-based encoder-decoder architecture. A special end-of-line token is introduced to stop the recurrent process. More recently, transformer-based architectures following the same principle have been proposed in \cite{kang2020,wick2021,TrOCR}. 

All these line-level architectures inherently require line-level segmentation annotations, which are very costly to produce. To alleviate this issue, some architectures handling single-column pages or paragraphs have been proposed recently. 
Yousef \textit{et al.} \cite{yousef2020} and Coquenet \textit{et al.} \cite{span} reformulate the two-dimensional problem as a one-dimensional problem in order to use the CTC loss, which was designed for one-dimensional sequence alignment problems only. Indeed, this loss enables one-shot predictions, contrary to the cross entropy loss, which implies a recurrent process. This way, prediction times are shorter. While the model from \cite{yousef2020} learns a representation transformation, the model from \cite{span} learns to align the predictions on the vertical axis, thanks to a reshaping operation.
\cite{Bluche2016} and \cite{van} proposed attention-based models. They are based on a recurrent implicit line-segmentation process. Compared to one-shot prediction approaches, the recurrent process implies longer prediction times, but it enables to model long term dependencies. Bluche \textit{et al.} used an MD-LSTM-based architecture. The model iterates a fixed number of times and all the probability lattices are concatenated and aligned at paragraph-level with the CTC loss. In \cite{van}, the model is an FCN+LSTM network that also learns to detect the end of the paragraph; alignment is performed line by line with the CTC loss.

Bluche \textit{et al.} \cite{bluche2017}, Singh \textit{et al.} \cite{singh2021}, and Rouhou \textit{et al.} \cite{Rouhoua2021} proposed different models which are based on an attention mechanism to predict the text character by character, trained with the cross-entropy loss and using a special end-of-transcription token.
While Bluche \textit{et al.} \cite{bluche2017} proposed an MD-LSTM model, a transformer-based model is used in \cite{singh2021,Rouhoua2021}. 
The models from \cite{singh2021} and \cite{Rouhoua2021} are trained to recognize, in addition to the text, the presence of non-textual areas (such as tables or drawings) or specific named entities (name, location for example), respectively.  These three works rely on curriculum learning using line or word segmentation annotations. Except for \cite{singh2021}, which uses a page-level private dataset, \cite{bluche2017} and \cite{Rouhoua2021} only evaluate their approach on paragraph images. 

As for the purely paragraph-level approaches, these attention models can be seen as a first step toward the integration of layout recognition through the text recognition process. As a matter of fact, the prediction of line break tokens implies the recognition of text line items. We propose to generalize this layout recognition to whole documents, integrating section and paragraph recognition, for instance.

Table \ref{tab:related_works} compares the different related works with the proposed approach. From left to right, the columns respectively denote the year and the reference, the handled task, the nature of the input, the context at decision time (global if the whole input signal is used, for attention mechanisms and recurrent layers for instance, or local otherwise), and the lowest level of physical segmentation annotation used, whether it is for training or pre-training.  

\begin{table*}[h!]
    \centering
    \caption{Comparison of the related works in terms of task, input, context, and physical segmentation annotation requirements.}
    \begin{tabular}{l l  c c c c }
       \multirow{2}{*}{Year} & \multirow{2}{*}{Reference}& \multirow{2}{*}{Task} & \multirow{2}{*}{Input} & \multirow{2}{*}{Context} & Segmentation \\
        & & & & & annotation\\
       \hline
       2016 & Voigtlaender \textit{et al.} \cite{Voigtlaender2016} & HTR & Line & Global & Line\\
       2016 & Bluche \textit{et al.} \cite{Bluche2016} & HTR & Paragraph & Global & Line\\
       2017 & Wigington \textit{et al.} \cite{Wigington2017} & HTR  & Line & Global & Line\\
       2017 & Bluche \textit{et al.} \cite{bluche2017}  & HTR & Paragraph & Global & Line\\
       2019 & Coquenet \textit{et al.} \cite{Coquenet2019}  & HTR  & Line & Local & Line\\
       2019 & Michael \textit{et al.} \cite{Michael2019} & HTR & Line & Global & Line\\
       2020 & Kang \textit{et al.} \cite{kang2020} & HTR  & Line & Global & Line\\
       2020 & Coquenet \textit{et al.} \cite{Coquenet2020} & HTR  & Line & Local & Line\\
       2020 & Yousef \textit{et al.} \cite{yousef2020} & HTR & Paragraph & Local & Paragraph\\
       2021 & Wick \textit{et al.} \cite{wick2021}  & HTR & Line & Global & Line\\
       2021 & Li \textit{et al.} \cite{TrOCR} & HTR & Line & Global & Line\\
       2021 & Coquenet \textit{et al.} \cite{span} & HTR & Paragraph & Local & Line\\
       2021 & Singh \textit{et al.} \cite{singh2021} & HTR + non-textual items & Paragraph & Global & Line\\
       2022 & Rouhou \textit{et al.} \cite{Rouhoua2021} & HTR + named entities & Paragraph & Global & Line\\
       2022 & Coquenet \textit{et al.} \cite{van} & HTR & Paragraph & Global & Line\\
       2022 & This work & HDR & Document & Global & \xmark\\ 
    \end{tabular}
    \label{tab:related_works}
\end{table*}

In this work, we propose the \modelacc{}, an end-to-end transformer-based model for whole handwritten document recognition, including the textual components and the logical layout information. This model is trained without using any segmentation label, and we evaluate it on two public datasets at page and double-page levels. To our knowledge, this is the first attempt that provides experimental results on such a task.

\section{Architecture and training}
\label{section-architecture}

The problem can be formalized as follows: the input is a raw document image $\mb{X}$ and the expected output is a sequence $\mseq{y}$ of tokens, of length $L_y$. Tokens are grouped in the same dictionary $\mset{D} = \mset{A} \cup \mset{S} \cup \{ \mathrm{<eot>} \}$, where $\mset{A}$ are tokens of characters from a given alphabet, $\mset{S}$ are specific layout tokens and <eot> is a special end-of-transcription token. Layout tokens are pairs of tokens (begin and end) that tag sequences of character tokens, as shown in Figure \ref{fig:xml}. As for characters, they vary according to the dataset used. In the following, we present the architecture and the training strategy.

\subsection{Architecture}
We propose the \modelname{} (\modelacc{}), an end-to-end encoder-decoder architecture that jointly recognizes both text and layout, from whole documents. We opted for an FCN as encoder since they are known to be efficient for feature extraction from images and can deal with inputs of variable sizes. For the decoder, we chose the transformer \cite{Vaswani2017} because it is currently the state-of-the-art approach for tasks involving the prediction of sequences of variable lengths.

\begin{figure}[ht!]
    \centering
    \includegraphics[width=\linewidth]{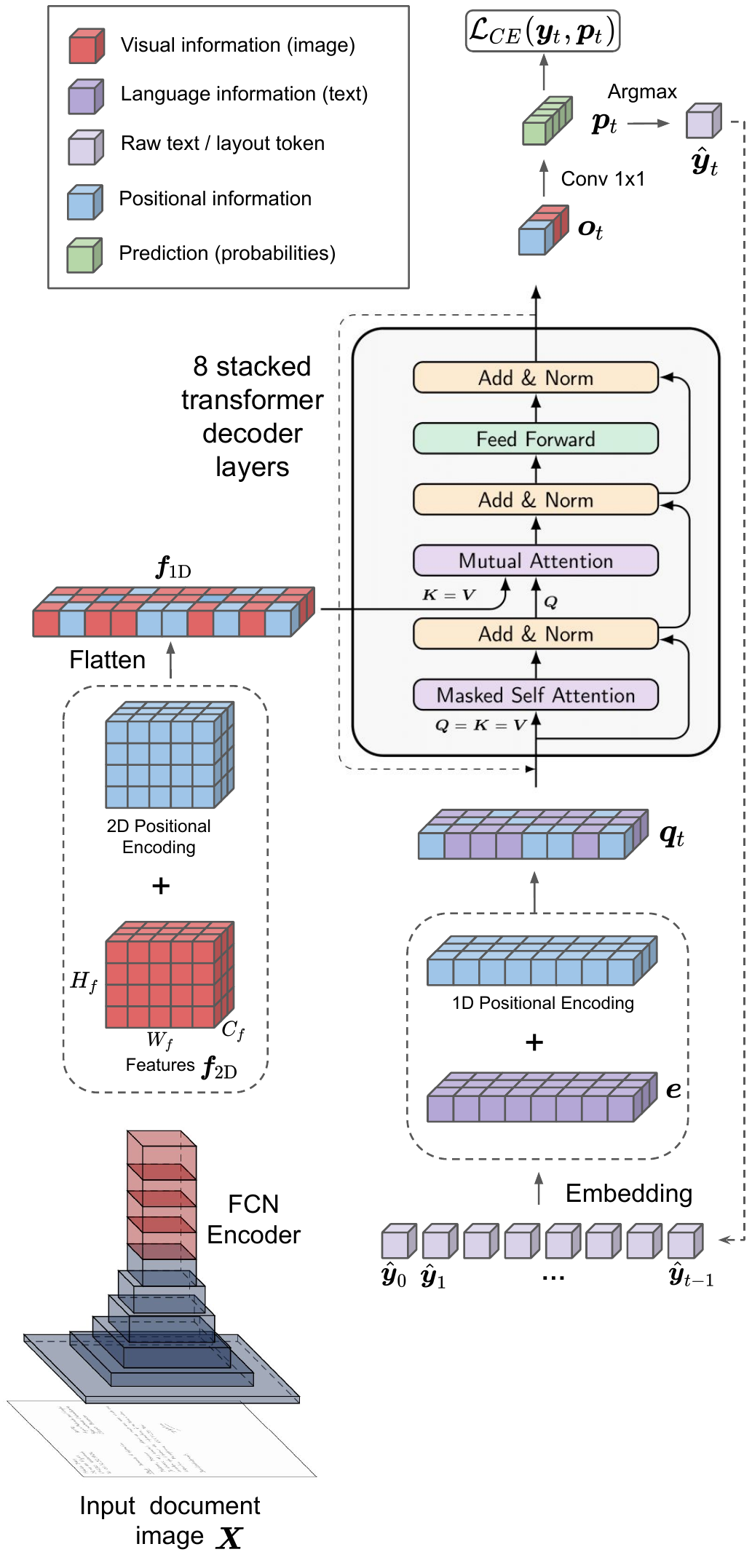}
    \caption{The \modelacc{} architecture is made up of an FCN encoder, for the extraction of 2D features $\mb{f}_\mathrm{2D}$, and a transformer-based decoder for the recurrent prediction of the character/layout tokens $\hat{\mseq{y}}_t$. At each iteration $t$, the model computes the representation $\mb{o}_t$ of the current character/layout token to recognize $\hat{\mseq{y}}_t$, based on the flattened features $\mb{f}_\mathrm{1D}$ and on the previous predictions. Positional encoding is added to these two modalities to preserve the spatial information through the transformer's attention mechanism.}
    \label{fig:model_overview}
\end{figure}

The \modelacc{} architecture is depicted in Figure \ref{fig:model_overview}. It is made up of an FCN encoder, to extracts 2D feature maps $\mb{f}_\mathrm{2D}$ from an input document image $\mb{X}$. 2D positional encoding is added to these features in order to keep the spatial information, before being flattened into a 1D sequence of features $\mb{f}_\mathrm{1D}$. This representation is computed only once and serves as input to the transformer decoder. The decoder follows a recurrent process at character-level: given the previously predicted tokens ($\hat{\mseq{y}}_0$, $\hat{\mseq{y}}_1$, \ldots, $\hat{\mseq{y}}_{t-1}$) and based on the computed features $\mb{f}_\mathrm{1D}$, it outputs the next token probabilities $\mb{p}_t$ for each token of $\mset{D}$. The final predicted token $\hat{\mseq{y}}_t$ is the one with the highest probability. The decoding process starts with an initial <sot> (start-of-transcription) token ($\hat{\mseq{y}}_0 = \mathrm{<sot>}$), and ends with the prediction of a special <eot> (end-of-transcription) token ($\hat{\mseq{y}}_{L_y+1} = \mathrm{<eot>}$).  We used the cross-entropy loss ($\mset{L}_\mathrm{CE}$) for training.

In the following sections, the encoder and the decoder are described in more detail.

\subsubsection{Encoder}
As proposed by Singh \textit{et al.} \cite{singh2021}, we opted for an FCN encoder in order to better model the local dependencies since inputs are images.
We used the FCN encoder of the Vertical Attention Network \cite{van} for many reasons. It achieves state-of-the-art results for HTR at paragraph level on many public datasets: RIMES 2011 \cite{RIMES}, IAM \cite{IAM} and READ 2016 \cite{READ2016}. It can handle inputs of variable sizes. And it implies few parameters (1.7M) compared to other approaches. 

The encoder takes as input a document image $\mb{X} \in \mathbb{R}^{H \times W \times C}$, with $H$, $W$ and $C$ being respectively the height, the width, and the number of channels ($C=3$ for an RGB image). It extracts some features maps for the whole document image: $\mb{f}_\mathrm{2D} \in \mathbb{R}^{H_f \times W_f \times C_f}$ with $H_f=\frac{H}{32}$, $W_f = \frac{W}{8}$ and $C_f = 256$.

The encoder is made up of a succession of 18 convolutional layers and 12 depthwise separable convolutional layers, with kernel $3 \times 3$ and ReLU activations. Diffused Mix Dropout \cite{van} and Instance Normalization are used to avoid overfitting and improve performance.

The original transformer architecture \cite{Vaswani2017} is defined for 1D sequences. Since the inputs are 2D images, we replaced the 1D positional encoding by 2D positional encoding, as proposed in \cite{singh2021}. 2D positional encoding is defined as a fixed encoding based on sine and cosine functions with different frequencies, in the same way as the positional encoding used by Vaswani \textit{et al.} \cite{Vaswani2017}; but instead of encoding a 1D position using all the channels, half is dedicated to vertical positional encoding and the other half to the horizontal positional encoding, as depicted in Equations \ref{eq:pos2d} and \ref{eq:pos2d2}:

\begin{equation}
\label{eq:pos2d}
\begin{split}
    \mathrm{PE}_\mathrm{2D}(x,y,2k) = \sin(w_k \cdot y), \\
    \mathrm{PE}_\mathrm{2D}(x,y,2k+1) = \cos(w_k \cdot y), \\
    \mathrm{PE}_\mathrm{2D}(x,y,d_{\mathrm{model}}/2 + 2k) = \sin(w_k \cdot x), \\
    \mathrm{PE}_\mathrm{2D}(x,y, d_{\mathrm{model}}/2 +2k+1) = \cos(w_k \cdot x), \\
    \forall k \in \left[0, d_\mathrm{model}/4\right],
\end{split}
\end{equation}
with 
\begin{equation}
\label{eq:pos2d2}
    w_k = 1/10000^{2k/d_\mathrm{model}}.
\end{equation}
We set $d_\mathrm{model}=C_f=256$.

Features $\mb{f}_\mathrm{2D}$ are summed with 2D positional encoding before being flattened for transformer decoder requirements following Equations \ref{eq:flatten} and \ref{eq:flatten2}:
\begin{equation}
\label{eq:flatten}
    \mb{f}_{\mathrm{1D}_j} = \mathrm{flatten}(\mb{f}_{\mathrm{2D}_{x,y}} + \mathrm{PE}_\mathrm{2D}(x,y)),
\end{equation}
with 
\begin{equation}
\label{eq:flatten2}
j = y W_f + x.
\end{equation}

\subsubsection{Decoder}
The decoder follows a recurrent process. At each iteration, it takes as input the flattened visual features $\mb{f}_\mathrm{1D}$ and the previously predicted tokens ($\hat{\mseq{y}}_0$, …, $\hat{\mseq{y}}_{t-1}$), and outputs the probabilities $\mb{p}_t$ for each token in the dictionary $\mset{D}$ at time step $t$. We used learned embeddings to convert the tokens to vectors $\mb{e}_{\hat{y}_i}$ of dimension $d_\mathrm{model}$. Embeddings are summed with 1D positional encoding corresponding to the position of the predicted tokens in the predicted sequence, as proposed by Vaswani \textit{et al.} for the original transformer \cite{Vaswani2017}:

\begin{equation}
    \mb{q}_{t,i} = \mathrm{PE}_\mathrm{1D}(i) + \mb{e}_{\hat{y}_i},
\end{equation}
with:
\begin{equation}
    \begin{split}
    \mathrm{PE}_\mathrm{1D}(x, 2k) = \sin(w_k \cdot x) \\
    \mathrm{PE}_\mathrm{1D}(x, 2k+1) = \cos(w_k \cdot x) \\
    \forall k \in \left[0, d_\mathrm{model}/2\right].
\end{split}
\end{equation}

The decoder is made up of a stack of 8 transformer decoder layers (as shown in Figure \ref{fig:model_overview}) followed by a convolutional layer with kernel $1 \times 1$ that computes the next token probabilities $\mb{p}_t$. The transformer decoder layers are based on multi-head attention \cite{Vaswani2017} mechanisms we denote as self-attention and mutual attention. Self-attention aims at modeling dependencies among the predicted sequence: it corresponds to multi-head attention where queries $Q$, keys $K$ and values $V$ are from the same input. Mutual attention is used to extract visual information from the encoder ($K$ and $V$ are from $\mb{f}_\mathrm{1D}$), based on $Q$ which comes from the previous predictions. In other words, given the previous predictions, it indicates where the model should look to predict the next token.

We used 8 decoder layers with dimension $d_\mathrm{model}$, feed forward dimension $d_\mathrm{model}$, 4 attention heads, ReLU activation and 10\% of dropout. Self attention is causal since it is based on the previous predictions. Following the approach proposed by Singh \textit{et al.} \cite{singh2021}, we used an attention window of length 100 for the self attention in order to reduce the computation time. It means that given an input sequence $\hat{\mseq{y}}$, the $t^\mathrm{th}$ output frame $\mb{o}_t$ is computed over the range $[\hat{\mseq{y}}_a, \hat{\mseq{y}}_{t-1}]$ with $a = \max(0, t-100)$.
The process starts with a <sos> token and ends when a <eot> token is predicted or after $L_{max}$ iterations. We set $L_{max}=3000$ to match the datasets needs.

The whole model is made up of 7.6 M trainable parameters and is trained in an end-to-end fashion using the cross-entropy loss over the sequence of tokens (of length $L_y$, and to which is added the special <eot> token):
\begin{equation}
    \mset{L} = \sum_{t=1}^{L_y+1} \mset{L}_\mathrm{CE}(\mseq{y}_t, \mb{p}_t)
\end{equation}

\subsection{Training strategy}
Training a deep attention-based neural network is difficult, especially when dealing with large inputs such as whole documents. The proposed training strategy is designed to improve the convergence with a limited amount of training data, and without using any segmentation label. It is performed in two steps:
\begin{itemize}
    \item Pre-training: the aim is to learn the feature extraction part of the \modelacc{}. We trained a line-level OCR model on synthetic printed lines and used it for transfer learning purposes for the \modelacc{}. We only used synthetic printed lines to avoid using segmentation labels, which are costly annotations to produce. Pre-training is carried out for 2 days with a mini-batch size of 16. Pre-processings, data augmentation and curriculum dropout are used during pre-training, as detailed afterwards.
    \item Training: the \modelacc{} is trained using teacher forcing (See section \ref{teacherforcing}) to reduce the training time per epoch. It is trained on both real and synthetic documents. The idea is to learn the attention mechanism, \textit{i.e.}, the reading order, through the synthetic images. Indeed, printed text is easier to recognize than handwritten text and the \modelacc{} is pre-trained on printed text lines. Once the reading order is learned, it becomes easier to adapt to real-world images. This is motivated by the nature of the reading order: it is the same between printed and handwritten documents sharing the same layout.
    
    Following this idea, the model is first trained with 90\% of synthetic documents during a curriculum phase to learn the reading order while using few real training samples. Then, this percentage is slowly decreased to reach 20\% through the epochs, in order to fine-tune on the real samples while keeping some synthetic samples acting as unseen training data.
    
    Training is carried out for 4 days. We did not use mini-batch: training is carried out image per image. Pre-processings, data augmentation, curriculum strategies and post-processings are used during training, as described in the following.
\end{itemize}

\subsubsection{Pre-training}
Training deep attention-based models is difficult. It is beneficial to train part of the model beforehand, whether it is the attention part or the feature extraction part, as shown in our previous works \cite{van}.
This way, we used a line-level pre-training strategy, \textit{i.e.}, we first train a line-level OCR model on isolated line images using the CTC loss. This line-level OCR model consists of an encoder (the same encoder as for the \modelacc{}) followed by an adaptive max pooling layer to collapse the vertical dimension, a convolutional layer and a softmax activation to predict the character and CTC blank label probabilities. However, contrary to \cite{van}, we do not use real isolated lines (extracted with the bounding boxes annotations) but synthetic printed text lines, generated from the text line transcriptions only. The \modelacc{} is then trained using transfer learning with this model to initialize the weights of the encoder and of the decision layer (last convolutional layer of the decoder) with those of this line-level OCR model.

\subsubsection{Curriculum strategies}
We used two curriculum strategies to improve the convergence by progressively increasing the difficulty of the task during training.

A curriculum strategy is used for the generation of synthetic data during the training of the \modelacc{}. Instead of directly generating whole documents, we progressively increase the number of lines per page contained in the generated documents. We set the minimum number of lines to 1 and the maximum number of lines to $l_\mathrm{max}$, to fit the properties of the datasets. In addition, we also crop the synthetic document image below the lowest text line during this curriculum stage. This way, we progressively increase both the length of the target sequence, through the number of text lines, and the input image size, through the iterations of the curriculum learning process. 

We used a second curriculum strategy regarding the dropout, as defined in \cite{Morerio2017}. It means that the dropout rate $\tau$ evolves during training:
\begin{equation}
    \tau_t = (1 - \bar{\tau}) \exp(- \frac{t}{T}) + \bar{\tau}, T > 0,
\end{equation}
where $\bar{\tau}$ is the final dropout rate, $t$ is the number of iterations (weight update) and $T$ is the total estimated number of weight updates during training. We set $T=5 \times 10^4$.

\subsubsection{Data Augmentation}
We used a data augmentation strategy with a probability of 90\%. This data augmentation strategy consists in applying some transformations, in random order, with a probability of 10\% for each one. These data augmentation techniques are: resolution modification, perspective transformation, elastic distortion, dilation, erosion, color jittering, Gaussian blur, Gaussian noise and sharpening. Data augmentation is applied to both synthetic and real images.

\subsubsection{Synthetic data}
We generated synthetic printed lines for pre-training and synthetic printed documents for training. We provide the related code \footnote{\url{https://github.com/FactoDeepLearning/DAN}}.
To this end, we arbitrarily chose a set of fonts $\mathcal{F}$ to introduce diversity in writing styles. We used these different fonts with various font sizes to bring more variability, making the model more robust. The original dataset $\mathcal{D}_\mathrm{doc}$ is used to extract isolated text line transcriptions $y_i$ associated to a layout class $c_i$, leading to a new dataset  $\mathcal{D}_\mathrm{line}$.
Synthetic lines are generated on the fly during pre-training, by randomly selecting a text line transcription from $\mathcal{D}_\mathrm{line}$.

While generating synthetic documents through learning has been studied in \cite{DocSyn} for instance, here we focused on a rule-based approach for simplicity. 
Algorithm \ref{alg:syn} details this generation process of synthetic documents. It is based on a style sheet $s$ which defines the different layout entities (classes) in the document and a set of constraints on them, such as relative and absolute positioning rules. It also defines properties for each layout entity: maximum height or width in pixels, characters per line, line width or number of lines. 
Synthetic documents are produced on the fly. A document image $\mb{X}$ is randomly chosen from the training dataset $\mathcal{D}_\mathrm{doc}$ to get a realistic document shape, which is used as a template for a synthetic document $\mb{D}$ to be generated. Given the current curriculum number of lines per page $l$ (between 1 and $l_\mathrm{max}$), the actual number of lines for the current synthetic document $l_\mathrm{doc}$ is randomly chosen between 1 and $l$. Layout entities are generated one after the other until reaching $l_\mathrm{doc}$: the layout class is chosen through "get\_next\_layout\_class" based on the style sheet definition and the current state of $\mb{D}$. Given the remaining number of lines ($l_\mathrm{doc}$ - $l_\mathrm{current}$), a random number of lines for the given entity $l_\mathrm{entity}$ is chosen in compliance with $s$. $l_\mathrm{entity}$ synthetic text line images are generated using a random font from $\mathcal{F}$ and a random text line from $\mathcal{D}_\mathrm{line}$. These images are concatenated on the vertical axis, introducing some random indent spacing. The associated ground truth transcriptions are also concatenated in the same order. The generated layout entity is then placed into $\mb{D}$ in compliance with $s$. Ground truth transcriptions of each layout entity are concatenated by adding the corresponding layout tokens, leading to the ground truth of the whole synthetic document $y$. The document image is cropped below the lowest layout entity, as part of the curriculum strategy.

\begin{algorithm*}
\caption{Synthetic document generation.}
\label{alg:syn}
\SetKwInOut{Input}{input}
\SetKwInOut{Output}{output}
\Input{
original document image $\mb{X}$, \\
number of lines $l_\mathrm{doc}$,\\
style sheet $s$,\\
line-level dataset $\mathcal{D}_\mathrm{line} = (\mathcal{Y},\mathcal{C})$,\\
set of fonts $\mathcal{F}$.
}
\Output{
synthetic document image $\mb{D}$,\\
ground truth $y$.
}
$y=$\textquote{ }\;
$l_\mathrm{current} = 0$\;
{\footnotesize \tcp{Get the size of a real image}}
$H, W = \mb{\mathrm{size}}(\mb{X})$\;
{\footnotesize \tcp{Initialize the synthetic document}}
$\mb{D} = \mb{\mathrm{zeros}}(H, W)$\;
{\footnotesize \tcp{While the document does not contain enough lines}}
\While{$l_\mathrm{current} < l_\mathrm{doc}$}{
    {\footnotesize \tcp{Get the class of the next layout entity to add, based on the style sheet}}
    $c = \mb{\mathrm{get\_next\_layout\_class}}(\mb{D}, s)$\;
    {\footnotesize \tcp{Randomly select the number of lines for this layout entity, constrained by the style sheet}}
    $l_\mathrm{entity} = \mb{\mathrm{get\_num\_lines}}(c, s, l_\mathrm{current}, l_\mathrm{doc})$ \;
    {\footnotesize \tcp{For each line of this layout entity}}
    \For{$k=1$ \KwTo $l_\mathrm{entity}$}{
        {\footnotesize \tcp{Randomly select a text line associated to this class}}
        $y_k = \mb{\mathrm{get\_random\_text}}(\mathcal{D}_\mathrm{line}, c)$ \;
        {\footnotesize \tcp{Randomly select a font and a size}}
        $f_k = \mb{\mathrm{get\_random\_font}}(\mathcal{F})$ \;
        {\footnotesize \tcp{Generate an image with the selected text and font}}
        $\mb{i}_k = \mb{\mathrm{generate\_text\_line\_image}}(y_k, f_k)$ \;
    }
    {\footnotesize \tcp{Concatenate the text line images of the current layout entity}}
    $\mb{i}_\mathrm{entity} = \mb{\mathrm{merge\_text\_line\_images}}(\mb{i}_1, ..., \mb{i}_{l_\mathrm{entity}})$ \;    
    {\footnotesize \tcp{Concatenate the character sequences of the corresponding text lines}}
    $y_\mathrm{entity} = \mb{\mathrm{merge\_text\_line\_gt}}(y_1, ..., y_{l_\mathrm{entity}})$ \;
    {\footnotesize \tcp{Add the layout entity image to the synthetic document, constrained by the style sheet}}
    $\mb{D} = \mb{\mathrm{add\_layout\_entity}}(\mb{D}, \mb{i}_\mathrm{entity}, s, c)$ \;
    {\footnotesize \tcp{Concatenate the character sequence of the layout entity to that of the synthetic document}}
    $y = \mb{\mathrm{add\_text}}(y, y_\mathrm{entity}, c)$ \;
    {\footnotesize \tcp{Update the number of lines in the synthetic document}}
    $l_\mathrm{current} += l_\mathrm{entity}$ \;
}
{\footnotesize \tcp{Crop the synthetic document image below the lowest text line}}
$\mb{D} = \mb{\mathrm{crop\_below\_lowest\_entity}}(\mb{D})$ \;
\end{algorithm*}

To sum up, we introduced variability in many points to generate different synthetic document examples:
\begin{itemize}
    \item different fonts and font sizes for the writing style.
    \item randomness and flexibility in the positioning constraints for the layout.
    \item random number of lines and mixed sample text lines for the content.
    \item cropping below the lowest text line for the image size.
\end{itemize}

\subsubsection{Teacher forcing}
\label{teacherforcing}
We used teacher forcing at training time to parallelize the computations by predicting the whole sequence at once: the ground truth is used in place of the previously predicted tokens. To make the \modelacc{} robust to errors occurring at prediction time, we introduced some errors in this sequence of pseudo previously predicted tokens. Some tokens are replaced by a random character or layout token. We found 20\% to be a good error rate through experiments.

\subsubsection{Pre-processings}
To reduce the memory consumption, images are downscaled through a bi-linear interpolation to a 150 dpi resolution. Images are normalized (zero mean, unit variance) based on mean and variance computed on the training set.

\subsubsection{Post-processings}
We used a rule-based post-processing to correct unpaired predicted layout tokens. This step is essential to compute the metrics. Let us denote <X> a layout start token and </X> its associated layout end token.
The post-processing consists of a forward pass on the whole predicted sequence $\hat{\mseq{y}}$ during which only tokens of layout are modified in order to have a well-formed global structure. The main rules are: 
\begin{itemize}
    \item a missing end token is added when there are two successive start tokens.
    \item isolated end tokens are removed. 
\end{itemize}
For instance, omitting text prediction for simplicity, the prediction "<X> <Y> </Y> </Z>" becomes "<X> </X> <Y> </Y>".

In addition, the post-processing ensures that the prediction is in accordance with the layout token grammar, \textit{i.e.}, the hierarchical relations between tokens are correct. It means that if a layout entity of class A can only be in an entity of class B, by definition, missing tokens are added. For example, the prediction "<A> </Y>" becomes "<B> <A> </A> </B>"
 
We used a second post-processing, for the text prediction: duplicated space characters are removed from the prediction.

\section{Metrics}
\label{section-metrics}
The proposed approach aims at jointly recognizing both text and layout. While there exist well-established metrics to measure the performance of both tasks independently, we are not aware of an adequate metric to evaluate both tasks when performed altogether.
To our knowledge, there is no prior work handling such a task. Therefore, we propose the evaluation of our approach using three different angles: the text recognition only, the layout recognition only and the joint recognition of both text and layout, using two new metrics named LOER and $\mathrm{mAP}_\mathrm{CER}$.

In the following, we will take as example the following predicted sequence $\hat{\mseq{y}}$, after post-processing:\\ 
"<X>text1</X><B><A>text2</A><A>text3</A></B>"

\subsection{Evaluation of the text recognition}
To evaluate the text recognition, all layout tokens are removed from the ground truth $\mseq{y}$ and from the prediction $\hat{\mseq{y}}$, leading to $\mseq{y}^\mathrm{text}$ and $\hat{\mseq{y}}^\mathrm{text}$, respectively. \\
Here, the above example becomes $\hat{\mseq{y}}^\mathrm{text}=$   "text1text2text3".

Then, we used the standard Character Error Rate (CER) and the Word Error Rate (WER) to evaluate the performance of the text recognition. 
They are both computed as the sum of the Levenshtein distances (noted $\mathrm{d}_\mathrm{lev}$) between the ground truths and the predictions, at document level, normalized by the total length of the ground truths $\mseq{y}_{\mathrm{len}_i}^\mathrm{text}$. For K examples in the dataset:
\begin{equation}
    \mathrm{CER} = \frac{\displaystyle \sum_{i=1}^K \mathrm{d_\mathrm{lev}}(\hat{\mseq{y}}_i^\mathrm{text}, \mseq{y}_i^\mathrm{text})}
                {\displaystyle\sum_{i=1}^K{\mseq{y}_{\mathrm{len}_i}^\mathrm{text}}}.
\end{equation}
WER is computed in the same way, but at word level. Punctuation characters are considered as words. 

One should note that, contrary to text line or paragraph recognition, the reading order is far more complicated to learn. An inversion in the reading order between two text blocks can severely impact the CER and WER values, even with correctly recognized text blocks.

\subsection{Evaluation of the layout recognition}
We cannot use existing DLA metrics, such as Intersection over Union (IoU), mean Average Precision (mAP) \cite{mAP_PascalVOC} or ZoneMap \cite{ZoneMap}, to evaluate the layout recognition, because they are based on physical layout (segmentation) annotations.

We decided to model the layout as an oriented graph to take into account both the reading order and the hierarchical relations between layout entities. To evaluate the layout recognition, we introduce a new metric: the Layout Ordering Error Rate (LOER). 
To this end, we associate to each ground truth and prediction a graph representation: $\mseq{y}^\mathrm{graph}$ and $\hat{\mseq{y}}^\mathrm{graph}$, as shown in Figure \ref{fig:dataset}. 
We propose to generate this graph representation in two steps. First, we compute $\mseq{y}^\mathrm{layout}$ and $\hat{\mseq{y}}^\mathrm{layout}$, the ground truth and the prediction from which all but layout tokens are removed. \\
Here, $\hat{\mseq{y}}^\mathrm{layout}=$ "<X></X><B><A></A><A></A></B>".\\
Second, we map this sequence of layout tokens into a graph following the hierarchical rules of the datasets. 

We designed the LOER following the same paradigm as CER, adapting it to graphs. It is computed as a Graph Edit Distance (GED), normalized by the number of nodes and edges in the ground truth. As shown in Figure \ref{fig:dataset}, this graph can be represented by ordering the nodes with respect to a root D which represents the document. This way, the graph can be represented as a multi-level graph where the nodes are the different layout entities, the oriented edges between successive levels (dashed arrows) are their hierarchy and the oriented edges inside the same level (solid arrows) represent their reading order. 

For K samples in the dataset, the LOER is computed as the sum of the graph edit distances,  normalized by the sum of the number of edges $n_\mathrm{e}$ and nodes $n_\mathrm{n}$ in the ground truths:
\begin{equation}
    \mathrm{LOER} = \frac{\displaystyle \sum_{i=1}^K\mathrm{GED}(\mb{y}^\mathrm{graph}_i, \hat{\mb{y}}^\mathrm{graph}_i)}{\displaystyle \sum_{i=1}^K n_{\mathrm{e}_i} + n_{\mathrm{n}_i}}.
\end{equation}
The graph edit distance is computed using a unit cost of edition whether it is for insertion, removal, or substitution and whether it is for edges or nodes. This computation becomes intractable in a reasonable running time for multiple pages. We circumvented this issue by assuming that the prediction of the page tokens was done in the right order. In this way, the GED of a document with several pages corresponds to the sum of the GED computed on the sub-graphs representing the isolated pages. Missing ground truth or prediction sub-graphs are compared to the null graph.

Combining CER and LOER is not sufficient to evaluate the correct recognition of the document. As a matter of fact, one could reach 0\% for both metrics by predicting first all the character tokens, and then all the layout tokens, each in the correct order. One misses the evaluation of the association between the layout tokens and their corresponding text parts.

\subsection{Evaluation of joint text and layout recognition}
We propose a second new metric, $\mathrm{mAP}_\mathrm{CER}$, to evaluate the joint recognition of both text and layout. It is based on the standard $\mathrm{mAP}$ score used for object detection approaches \cite{mAP_COCO,mAP_PascalVOC}; but instead of using the IoU to consider a prediction as true or false, we use the CER. It is computed as follows:
\begin{itemize}
    \item The predicted sequence $\hat{\mseq{y}}$ and the ground truth sequence $\mseq{y}$ are split into sub-sequences. These sub-sequences are extracted using the start and end tokens of a same class. Then, they are grouped by classes into lists of sub-sequences. Results for our prediction example are given in Table \ref{tab:mAP}.
    
    For each layout class, sub-sequences are ordered by their confidence score, computed as the mean between prediction probabilities associated with the start and end tokens of this class (probabilities from $\mseq{p}_t$). A predicted sub-sequence is considered as true positive if the CER between this sub-sequence and a sub-sequence of the ground truth from the same class is under a given threshold. Otherwise, it is considered as false positive. When associated, the used sub-sequences are removed until there is no more sub-sequence in the ground truth or in the prediction.
    
    \begin{table}[h]
     \caption{Sub-sequences are extracted, grouped and ordered by layout classes for $\mathrm{mAP}_\mathrm{CER}$ computation. Left: tokens of the predicted sequence and associated confidence score. Consecutive text tokens are grouped for simplicity, their associated confidence score has been averaged. Right: sub-sequences are extracted by layout tokens and ordered given confidence score.}
     \label{tab:mAP}
        \hfill
        \begin{subfloat}
            \centering
            \resizebox{0.15\textwidth}{!}{
            \begin{tabular}{c c}
            Token & Confidence\\
            \hline
            <X> & 90\% \\
            text1 & 95\% \\
            </X> & 70\% \\
            <B> & 95\% \\
            <A> & 82\% \\
            text2 & 73\% \\
            </A> & 86\% \\
            <A> & 80\% \\
            text3 & 89\% \\
            </A> & 80\% \\
            </B> & 75\% \\
           \end{tabular}
           }
        \end{subfloat}
        \hfill
        \begin{subfloat}
            \centering
            \resizebox{0.17\textwidth}{!}{
            \begin{tabular}{c c}
            Score & Text \\
            \hline
            \multicolumn{2}{l}{\textbf{X}} \\
            80\% & text1\\
            \multicolumn{2}{l}{\textbf{A}} \\
            84\% & text2\\
            80\% & text3\\
            \multicolumn{2}{l}{\textbf{B}} \\
            85\% & text2text3\\

            \end{tabular}
            }
        \end{subfloat}
        
        \hfill
\end{table}
    
    \item The average precision $\mathrm{AP}_c$ for a layout class $c$ corresponds to the Area Under the precision-recall Curve (AUC). Based on the Pascal VOC challenge \cite{mAP_PascalVOC}, it is computed as an approximation by summing the rectangular areas under the curve, formed by each modification of the precision $p_n$ and recall $r_n$:
    \begin{equation}
        \mathrm{AP}_{\mathrm{CER}_c} = \sum (r_{n+1} - r_n) \cdot p_\mathrm{interp}(r_{n+1}),
    \end{equation}
    with 
    \begin{equation}
        p_\mathrm{interp}(r_{n+1}) = \max_{\tilde{r}>r_{n+1}} p(\tilde{r}).
    \end{equation}
    
    \item The average precision is itself averaged for different CER thresholds, between $\theta_\mathrm{min} = 5\%$ and $\theta_\mathrm{max} = 50\%$ with a step $\Delta\theta = 5\%$, leading to 10 thresholds:
    \begin{equation}
        \mathrm{AP}_{\mathrm{CER}_c}^{5:50:5} = \frac{1}{10} \displaystyle \sum_{k=1}^{10}\mathrm{AP}_{\mathrm{CER}_c}^{5k}.
    \end{equation}
    
    \item The mean average precision for a document is then computed as a weighted sum over the different layout classes, weighted by the number of characters $\mathrm{len}_c$ in each class $c$:
    \begin{equation}
        \mathrm{mAP}_\mathrm{CER} = \frac{\displaystyle \sum_{c \in S} \mathrm{AP}_{\mathrm{CER}_c}^{5:50:5} \cdot \mathrm{len}_c }{\displaystyle \sum_{c \in S}  \mathrm{len}_c}
    \end{equation}
    
    \item Finally, the global mAP for a set of documents is computed by averaging the mAP of the different documents, weighted by the number of characters in each document.
\end{itemize}

This way, the $\mathrm{mAP}_\mathrm{CER}$ gives an idea of how well the text regions have been classified, based on the recognized text. It could not be used with the CER only because it does not evaluate the order of the classified text regions.  Combining CER, LOER and $\mathrm{mAP}_\mathrm{CER}$ enables to have a real estimation of the quality of the prediction for joint text and layout recognition.

However, one could argue that the layout recognition performance is biased by the post-processing we used. To evaluate the impact of the post-processing in the final results, we defined a dedicated metric: the Post-Processing Edition Rate (PPER). It is used to understand how much of the layout recognition is due to the raw network prediction and how much is due to the post-processing. It is defined by the number of post-processing edition operations (addition or removal of layout tokens) $n_\mathrm{ppe}$, normalized by the number of layout tokens in the ground truth $\mb{y}^{\mathrm{layout}}_\mathrm{len}$. For $K$ examples:
\begin{equation}
    \mathrm{PPER} = \frac{ \sum_{i=1}^{K} n_{\mathrm{ppe}_i}}{\sum_{i=1}^{K} \mb{y}^{\mathrm{layout}}_{\mathrm{len}_i}}.
\end{equation}
One should keep in mind that this metric only quantifies how much of the final layout prediction is due to post-processing through edition operations: these modifications can be either beneficial or unfavorable.
\section{Experiments}
\label{section-experiments}
This section is dedicated to the evaluation of the \modelname{} (\modelacc{}) for document recognition. We evaluate the \modelacc{} on the RIMES 2009 and READ 2016 datasets. We provide a visualization of the attention process and of the predictions. We also provide an ablation study to highlight the key components that made it possible to achieve these results.

\subsection{Datasets}
We evaluated the \modelacc{} on two public handwritten document datasets: RIMES \cite{RIMES_page,RIMES} and READ 2016 \cite{READ2016}. We also evaluate the \modelacc{} on the MAURDOR dataset \cite{MAURDOR} for the text recognition task only, as detailed in the appendix.

\subsubsection{RIMES}
The RIMES dataset corresponds to French gray-scale handwritten page images at a resolution of 300 dpi. They have been produced in the context of writing mail scenarios. We used the page-level version of the dataset used during the 2009 evaluation campaign \cite{RIMES_page}, referred to as RIMES 2009. It is split into 1,050 pages for training, 100 pages for validation and 100 pages for test. We used two kinds of annotation from this page-level dataset: transcription ground truth and layout analysis annotations. Text regions are classified among 7 classes: sender (S), recipient (R), date \& location (W), subject (Y), opening (O), body (B) and PS \& attachment (P). We used these classes and associated them with the corresponding text part: we do not use any positional ground truth to train the proposed model. We corrected 3 annotations where at least one text line was missing. Sometimes, the annotators proposed two options for a given word or expression, we systematically selected the first one in every case. 

Since there is no other work evaluating their model on this dataset, we also evaluate the performance of the \modelacc{} on the RIMES 2009 dataset at paragraph-level, as well as on the RIMES 2011 dataset \cite{RIMES} at paragraph and line levels. The idea is to compare the recognition performance between the different segmentation levels and with the state-of-the-art approaches which rely on pre-segmented input at line or paragraph level. One should notice that the paragraphs from RIMES 2011 are not extracted from RIMES 2009, so we cannot directly compare the results on both datasets, but it is the nearest comparison we can make.

\subsubsection{READ 2016}
The READ 2016 dataset is a subset of the Ratsprotokolle collection (READ project). It was proposed for the ICFHR 2016 competition on handwritten text recognition and corresponds to Early Modern German handwritings. We used the page-level version of the dataset. We also generated a double-page version of this same dataset by concatenating the images of successive pages. Only few pages of the original dataset are not paired and have been removed. Based on their positions, we automatically added a class to each text region among the 5 following classes: page (P), page number (N), body (B), annotation (A) and section (S) (group of linked annotation(s) and body). For comparison purposes, we also evaluate the \modelacc{} on the READ 2016 dataset at line and paragraph levels.

\begin{figure*}[h!]
\centering
    \begin{subfigure}{0.49\textwidth}
    \centering
    \includegraphics[width=\textwidth]{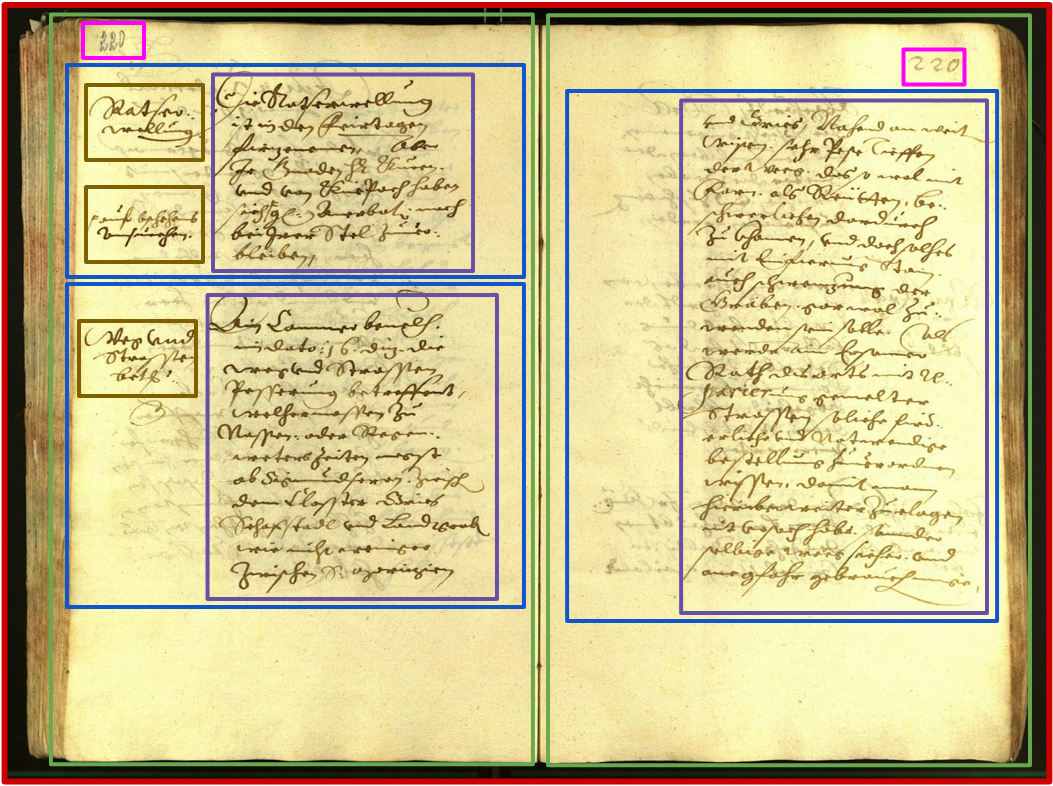}   
    \par\medskip
    \includegraphics[width=0.7\textwidth]{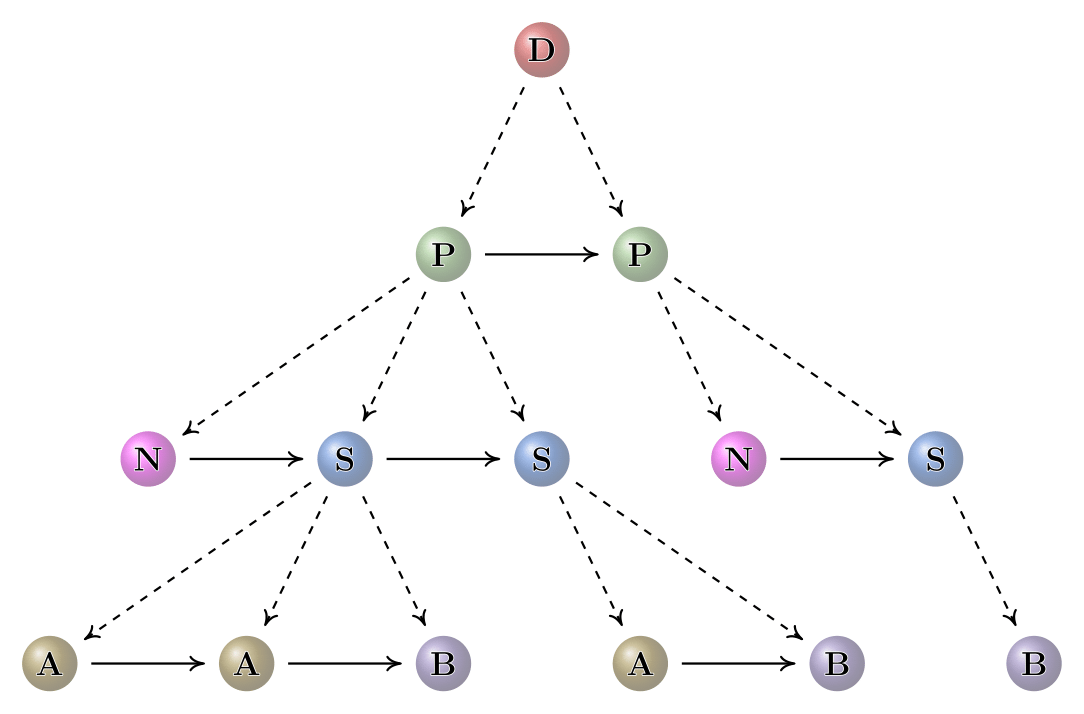}   
    \end{subfigure}
    \hfill
    \begin{subfigure}{0.49\textwidth}
    \centering
    \includegraphics[width=0.7\textwidth]{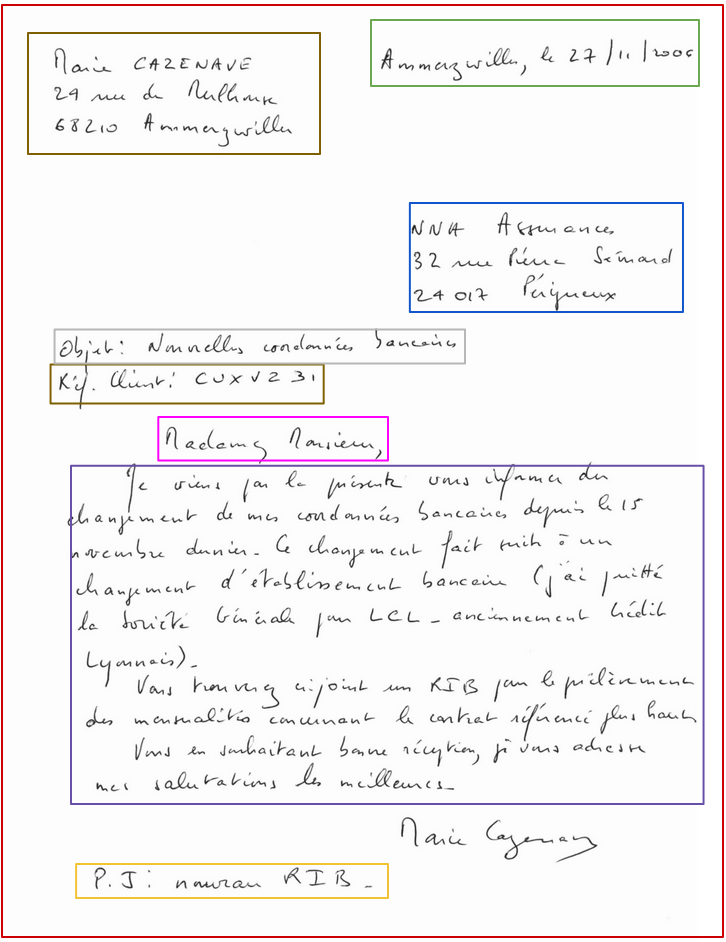}   
    \par\medskip 
    \includegraphics[width=0.9\textwidth]{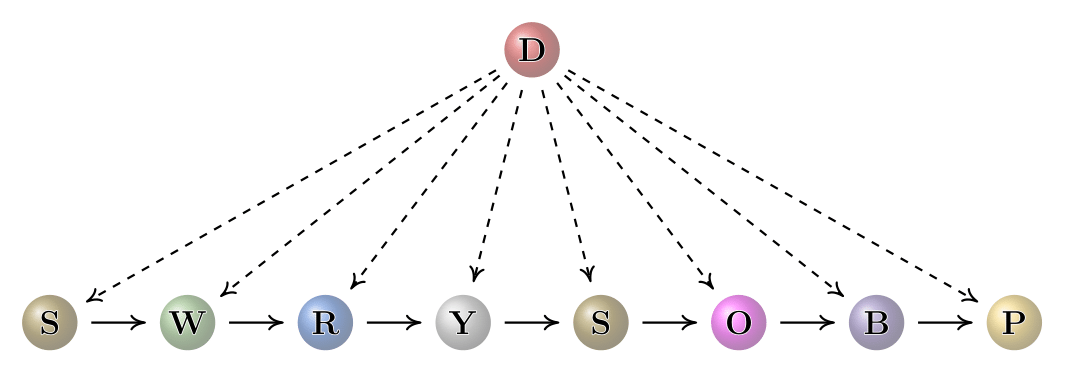}  
    \end{subfigure}

    \caption{Images from READ 2016 and RIMES 2009 and associated layout graph annotation.}
\label{fig:dataset}
\end{figure*}

For both RIMES 2009 and READ 2016 datasets at page or double-page level, the reading order is deduced automatically from the paragraph positions as follows. 
For READ 2016, the reading order is from page to page: first, the page number, then section by section. Among a section, all annotations are read before the body.
For RIMES 2009, text regions are read from top to bottom. If two text regions share the same vertical space, text regions are read from left to right.
An example of each dataset is depicted in Figure \ref{fig:dataset}. Text regions are represented by bounding boxes, colored given their class. We also represented the expected reading order as an oriented graph linking the different text regions. Both datasets have their own specificity. READ 2016 has a more regular layout compared to RIMES, but it includes hierarchical layout tokens: bodies are included in sections, themselves included in pages. In addition, it can contain multiple pages. RIMES 2009 provides more variability regarding the layout, but the layout tokens are sequential.

Datasets are split into training, validation and test sets, as detailed in Table \ref{table:split}. It corresponds to the official splits, or the most used by the community. We also provide the number of characters for each dataset, as well as the number of layout tokens (two by class: begin and end). We also provide this information at line and paragraph levels for comparison purposes.

\begin{table}[h!]
    \caption{Datasets split in training, validation and test sets and associated number of tokens in their alphabet.}
    \centering
    \resizebox{\linewidth}{!}{
    \begin{tabular}{ c c c c c c c}
    \hline
    \multirow{2}{*}{Dataset} & \multirow{2}{*}{Level} & \multirow{2}{*}{Training} & \multirow{2}{*}{Validation} & \multirow{2}{*}{Test} & \# char  & \# layout\\ 
     & & & & & tokens & tokens\\
    \hline
    \hline     
    \multirow{2}{*}{RIMES 2011} & Line & 10,530 & 801 &  778 & 97 & \xmark\\
    & Paragraph & 1,400 & 100 & 100 & 98 & \xmark\\ 
    \hline  
    \multirow{2}{*}{RIMES 2009} & Paragraph & 5,875 & 540 &  559 & 108 & \xmark\\
    & Page & 1,050 & 100 & 100 & 108 & 14\\
    \hline  
    \multirow{4}{*}{READ 2016} & Line & 8,367 & 1,043 & 1,140 & 88 & \xmark\\
    & Paragraph & 1,602 & 182 & 199 & 89 & \xmark\\
    & Page & 350 & 50 & 50 & 89 & 10\\
    & Double page & 169 & 24 & 24 & 89 & 10\\

    \hline
    \end{tabular}
    }
    \label{table:split}
\end{table}

\subsection{Training details}
Pre-training and training are carried out with the same following configuration, no matter the dataset:
\begin{itemize}
    \item PyTorch framework with Automatic Mixed Precision.
    \item Training with a single GPU Tesla V100 (32 Gb).
    \item Adam optimizer with an initial learning rate of $10^{-4}$.
    \item We use exactly the same hyperparameters for both datasets.
    \item We do not use any external data, external language model nor lexicon constraints.
\end{itemize}

\begin{figure}[ht]
    \centering
    \begin{subfigure}[b]{\linewidth}
    {
    \setlength{\fboxsep}{0pt}%
    \setlength{\fboxrule}{1pt}%
    \fbox{\includegraphics[width=\linewidth]{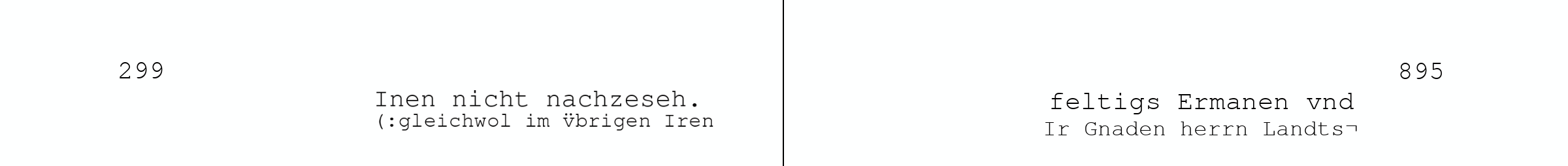}}%
    }
    \caption{$l=3$.}
    \end{subfigure}
    \par\medskip
    \begin{subfigure}[b]{\linewidth}
        {
    \setlength{\fboxsep}{0pt}%
    \setlength{\fboxrule}{1pt}%
    \fbox{\includegraphics[width=\linewidth]{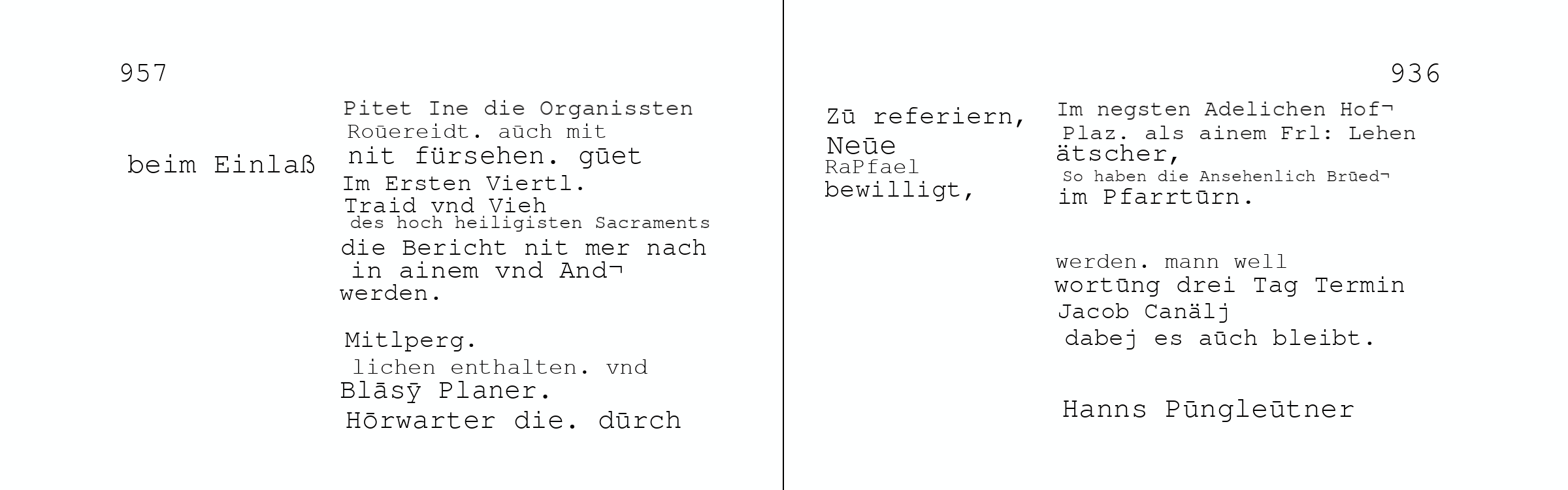}}%
    }
    \caption{$l=15$.}
    \end{subfigure}
    \par\medskip
    \begin{subfigure}[b]{\linewidth}
        {
    \setlength{\fboxsep}{0pt}%
    \setlength{\fboxrule}{1pt}%
    \fbox{\includegraphics[width=\linewidth]{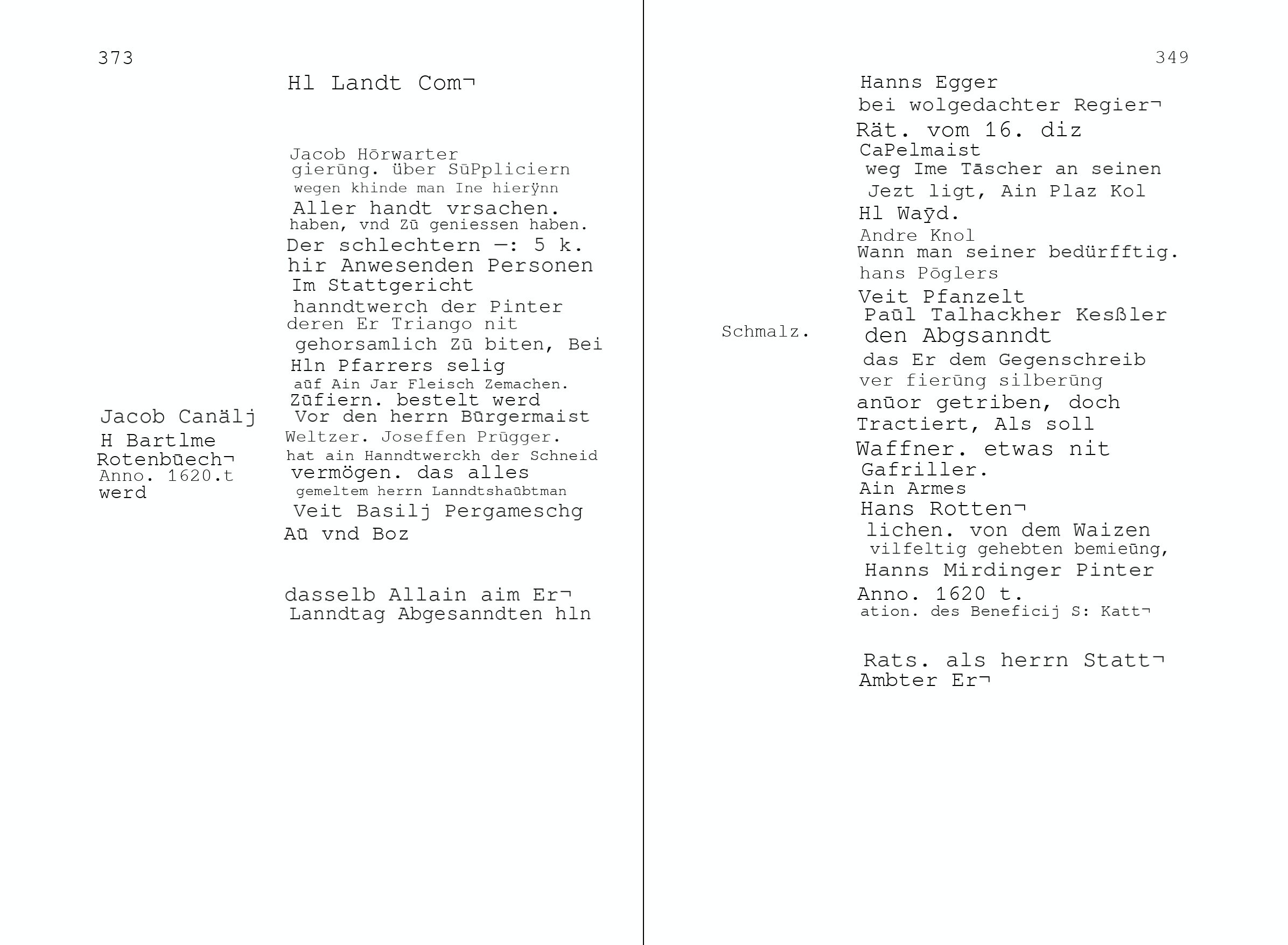}}%
    }
    \caption{$l=l_\mathrm{max}=30$ (end of curriculum stage, no crop).}
    \end{subfigure}
    
    \caption{Illustration of the curriculum learning strategy through the synthetic document image generation process for the READ 2016 dataset at double-page level. The number of lines per page $l$ increases from 1 to 30 through the epochs. The cropping strategy is discarded when the curriculum phase ends.}
    \label{fig:syn}
\end{figure}

For the generation of synthetic documents, we set the maximum number of lines per page $l_\mathrm{max}$ to 30 for READ 2016 and to 40 for RIMES 2009 to match the dataset properties. Given a set of arbitrarily-chosen fonts, we only kept those for which all characters were supported, leading to 41 fonts for READ 2016 and 95 for RIMES 2009. The font lists are provided with the code for reproducibility purposes. Figure \ref{fig:syn} illustrates the curriculum process for the generation of synthetic documents for the READ 2016 dataset at double-page level. As one can note, these synthetic documents are far from being visually realistic compared to the original dataset. This does not matter, since the objective here is only to learn the reading order.

We also evaluate the \modelacc{} at paragraph and line levels for comparison purposes. For each training of the same dataset, at paragraph and page levels of RIMES 2009 for instance, the same pre-trained weights are used to initialize the model. Each evaluation corresponds to specific training. Evaluation at paragraph or line levels corresponds to training only on synthetic and real paragraphs or lines, respectively. 

\subsection{Evaluation}

To our knowledge, there is no work evaluating their system on the RIMES 2009 and READ 2016 datasets at page level. Comparison at paragraph and line levels is carried out with approaches under similar conditions, \textit{i.e.}, without external data nor external language model. In Table \ref{table:rimes}, we present the evaluation of the \modelacc{} on the RIMES 2009 dataset at paragraph and page levels. One can notice that we reach very satisfying results at page level for both text and layout recognition with a CER of 4.54\%, a WER of 11.85\%, a LOER of 3.82\% and a $\mathrm{mAP}_\mathrm{CER}$ of 93.74\%.
The closest dataset with which we can compare is the RIMES 2011 dataset at paragraph level \cite{RIMES}. The \modelacc{} achieves new state-of-the-art results at paragraph level and competitive results at line level on this dataset. One should keep in mind that, as said previously, the comparison between RIMES 2009 and RIMES 2011 at paragraph level is not fair because RIMES 2011 only contains body images whose content seems easier to recognize than that of RIMES 2009. Unique character sequences representing dates, postal codes, product and client references, or even proper nouns like names and places, are mainly in the other text regions. 
In addition, the body images from the RIMES 2011 dataset are not taken from the page images of RIMES 2009. This explains the CER difference between RIMES 2009 (5.46\%) and RIMES 2011 (1.82\%) at paragraph level. One can notice a CER improvement from line to paragraph level (for RIMES 2011) and from paragraph to page level (for RIMES 2009). This highlights the negative impact of using a prior segmentation step, which is prone to annotation variations or errors.

\begin{table*}[ht]
    \caption{Evaluation of the DAN on the test set of the RIMES datasets and comparison with the state-of-the-art approaches.}
    \centering
    \resizebox{0.8\linewidth}{!}{
    \begin{threeparttable}[b]
        \begin{tabular}{ l l c c c c c}
        \hline
        Dataset & Approach & CER $\downarrow$ & WER $\downarrow$ & LOER $\downarrow$ & $\mathrm{mAP}_\mathrm{CER}$ $\uparrow$ & PPER $\downarrow$\\ 
        \hline
        \hline
        \multirow{9}{*}{\shortstack{RIMES\\2011\\\cite{RIMES}}}& \textbf{Line level}\\
        & Coquenet \textit{et al.} \cite{van} (FCN) & 3.04\% & 8.32\% & \xmark & \xmark & \xmark \\
        & Puigcerver \textit{et al.} \cite{Puigcerver2017} (CNN+BLSTM\tnote{a} ) & \textbf{2.3\%} & 9.6\% & \xmark & \xmark & \xmark\\
        & Ours (DAN\tnote{c} ) & 2.63\% & \textbf{6.78\%} & \xmark & \xmark & \xmark\\
        & \textbf{Paragraph level}\\
        & Coquenet \textit{et al.} \cite{span} (FCN) & 4.17\% & 15.61\% & \xmark & \xmark & \xmark \\
        & Bluche \textit{et al.} \cite{Bluche2016} (CNN+MDLSTM\tnote{b} ) & 2.9\% & 12.6\% & \xmark & \xmark & \xmark \\
        & Coquenet \textit{et al.} \cite{van} (FCN+LSTM\tnote{b} )  & 1.91\% & 6.72\% & \xmark & \xmark & \xmark \\
        & Ours (DAN\tnote{c} ) & \textbf{1.82\%} & \textbf{5.03\%} & \xmark & \xmark & \xmark\\
        \hline
        \multirow{4}{*}{\shortstack{RIMES\\2009\\\cite{RIMES_page}}} & \textbf{Paragraph level}\\
        & Ours (DAN\tnote{c} ) & 5.46\% & 13.04\% & \xmark & \xmark & \xmark \\
        & \textbf{Page level}\\
        & Ours (DAN\tnote{c} ) & 4.54\% & 11.85\% & 3.82\% & 93.74\% & 1.45\%\\
        \hline
        \end{tabular}
        \begin{tablenotes}
            \item [a] This work uses a different split (10,203 for training, 1,130 for validation and 778 for test).
            \item [b] with line-level attention.
            \item [c] with character-level attention.
        \end{tablenotes}
    \end{threeparttable}
    }
    \label{table:rimes}
\end{table*}

Table \ref{table:read} provides an evaluation of the \modelacc{} on the READ 2016 dataset, at line, paragraph, single-page and double-page levels. As one can note, we achieve state-of-the-art results at line and paragraph levels. We also reach very interesting results whether it is at single-page or double-page level, with a CER of 3.43\% and 3.70\%, respectively. It corresponds to slightly higher CER compared to the paragraph level (3.22\%).
One can note that the LOER and the $\mathrm{mAP}_\mathrm{CER}$ are also satisfying, highlighting the good recognition of the layout. Moreover, these metrics are slightly better for the double-page level dataset. This could be explained by the higher necessity to understand the layout for complex samples. This is discussed in Section \ref{section:ablation}.

\begin{table*}[ht]
    \caption{Evaluation of the DAN on the test set of the READ 2016 dataset and comparison with the state-of-the-art approaches}
    \centering
    \resizebox{0.7\linewidth}{!}{
    \begin{threeparttable}[b]
        \begin{tabular}{ l c c c c c}
        \hline
        Approach & CER $\downarrow$ & WER $\downarrow$ & LOER $\downarrow$ & $\mathrm{mAP}_\mathrm{CER}$ $\uparrow$ & PPER $\downarrow$\\ 
        \hline
        \hline
        \textbf{Line level}\\
        Michael \textit{et al.} \cite{Michael2019} (CNN+BLSTM\tnote{a} )  & 4.66\% & \xmark  & \xmark & \xmark & \xmark \\
        Sánchez \textit{et al.} \cite{READ2016} (CNN+RNN) & 5.1\% & 21.1\% & \xmark & \xmark & \xmark\\
         Coquenet \textit{et al.} \cite{van} (FCN+LSTM\tnote{b} )  & \textbf{4.10\%} & \textbf{16.29\%} & \xmark & \xmark & \xmark\\
        Ours (DAN\tnote{a} ) & \textbf{4.10\%} & 17.64\% & \xmark & \xmark & \xmark\\
        \textbf{Paragraph level}\\
         Coquenet \textit{et al.} \cite{span} (FCN) & 6.20\% & 25.69\% & \xmark & \xmark & \xmark\\
         Coquenet \textit{et al.} \cite{van} (FCN+LSTM\tnote{b} )  & 3.59\% & 13.94\% & \xmark & \xmark & \xmark\\
        Ours (DAN\tnote{a} ) & \textbf{3.22\%} & \textbf{13.63\%} & \xmark & \xmark & \xmark\\
        \textbf{Single-page level}\\
        Ours (DAN\tnote{a} ) & 3.43\% & 13.05\% & 5.17\% & 93.32\% & 0.14\%\\
        \textbf{Double-page level}\\
        Ours (DAN\tnote{a} ) & 3.70\% & 14.15\% & 4.98\% & 93.09\% & 0.15\%\\
        \hline
        \end{tabular}
        \begin{tablenotes}
            \item [a] with character-level attention.
            \item [b] with line-level attention.
        \end{tablenotes}
    \end{threeparttable}
    }
    \label{table:read}
\end{table*}

One should note that these CER values, for both RIMES 2009 and READ 2016 datasets, should not be compared directly to paragraph-level or line-level HTR approaches. Indeed, it is necessary to understand the impact of the reading order. CER is computed based on the edit distance between two one-dimensional sequences. This way, if the reading order is wrong, it impacts severely the CER. For instance, if the sender coordinates are read before the recipient coordinates while it is the opposite in the ground truth, the edit distance will be very important even if the text is well recognized. It means that part of this CER is due to a wrong reading order and not to wrong text recognition.
However, $\mathrm{mAP}_\mathrm{CER}$ would not be impacted: it is invariant to the reading order between the different text regions.

For both datasets, the PPER metric is very low. It indicates that the good results obtained for the layout recognition are mainly due to the \modelacc{} itself and not to the post-processing stage.

We now focus on the $\mathrm{mAP}_\mathrm{CER}$ metric with the READ 2016 dataset at double-page level. In Table \ref{tab:map_read}, we detailed this metric for each layout class and for each threshold of CER. As one can notice, pages, page numbers, sections, and bodies are very well recognized with at least 91\% for a CER as of 10\%. The \modelacc{} has more difficulty with the annotations, with an average of 80.28\%. We assume that this is due to two main points: it is the layout entity with the most variability. There can be zero, one or multiple annotations per body. In addition, they can be placed wherever along the body, from its beginning to its end. The second point is about the length of the annotations: they are very much shorter than bodies. This way, only few errors can lead to an important CER increase, leading to lower average precision.
\begin{table*}[ht]
    \caption{$\mathrm{mAP}_\mathrm{CER}$ detailed for each class and each CER threshold, for the READ 2016 double-page dataset.}
    \centering
    \resizebox{\linewidth}{!}{
    \begin{tabular}{ l c c c c c c c c c c c}
    \hline
    & 
    $\mathrm{AP}_\mathrm{CER}^{5:50:5}$ & $\mathrm{AP}_\mathrm{CER}^{5}$ & $\mathrm{AP}_\mathrm{CER}^{10}$ & $\mathrm{AP}_\mathrm{CER}^{15}$ &
    $\mathrm{AP}_\mathrm{CER}^{20}$ & $\mathrm{AP}_\mathrm{CER}^{25}$ & $\mathrm{AP}_\mathrm{CER}^{30}$ &
    $\mathrm{AP}_\mathrm{CER}^{35}$ & $\mathrm{AP}_\mathrm{CER}^{40}$ & $\mathrm{AP}_\mathrm{CER}^{45}$ & $\mathrm{AP}_\mathrm{CER}^{50}$\\
    \hline
    \hline
    Page (P) & 96.56 & 71.88 & 93.75 & 100.00 & 100.00 & 100.00 & 100.00 & 100.00 & 100.00 & 100.00 & 100.00\\
    Page number (N) & 97.50 & 95.83 & 95.83 & 95.83 & 95.83 & 95.83 & 95.83 & 100.00 & 100.00 & 100.00 & 100.00 \\
    Section (S) & 94.04 & 73.31 & 91.00 & 94.70 & 96.48 & 96.48 & 97.46 & 97.46 & 97.46 & 97.46 & 98.57 \\
    Annotation (A) & 80.28 & 37.84 & 60.07 & 75.49 & 88.24 & 88.24 & 90.20 & 90.20 & 90.20 & 90.20 & 92.16 \\
    Body (B) & 95.18 & 85.74 & 93.33 & 95.93 & 95.93 & 95.93 & 96.98 & 96.98 & 96.98 & 96.98 & 96.98\\
    \hline
    \end{tabular}
    }
    \label{tab:map_read}
\end{table*}

The average inference time per image is given in Table \ref{table:inference} for both RIMES 2009 and READ 2016 datasets. As one can note, the inference time grows linearly with the number of characters: 5.8s for an average of 588 characters for RIMES 2009, 4.3s for 468 characters for READ 2016 at single-page level and 9.7s for the double-page level. It has to be noted that the input size does not seem to have a significant impact on the inference time, which is mainly due to the recurrent, attention-based decoding process.
\begin{table}[h!]
    \caption{Prediction using a single GPU V100 (32 Gb). Times and properties are averaged for a sample of the test set.}
    \centering
    \resizebox{\linewidth}{!}{
    \begin{tabular}{ l | c c c c }
    \hline
    Dataset & Time & Input size & \# lines & \# chars \\
    \hline
    \hline
    RIMES 2009 (page) & 5.8s & $3{,}502 \times 2{,}471$ & 18 & 588\\
    READ 2016 (single page) & 4.3s & $3{,}510 \times 2{,}380$ & 23 & 468\\
    READ 2016 (double-page) & 9.7s & $3{,}510 \times 4{,}760$ & 46 & 944\\ 
    \hline
    \end{tabular}
    }
    \label{table:inference}
\end{table}

\subsection{Visualization}
\begin{figure*}[ht]
    \centering
    \includegraphics[width=0.7\textwidth]{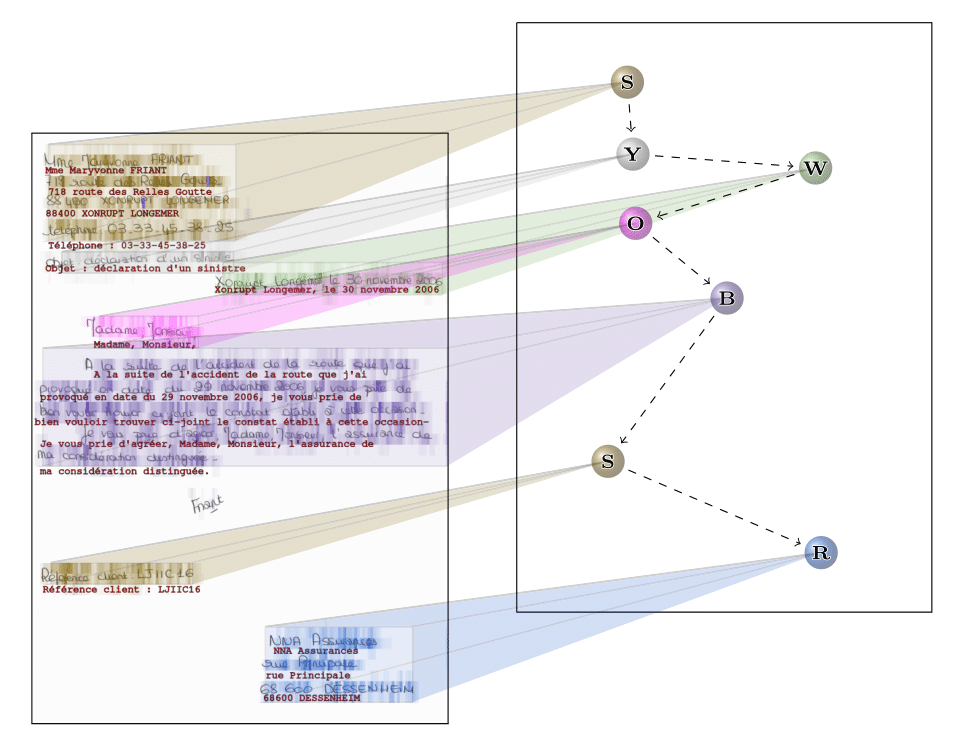}
    \caption{Visualization of the prediction on a RIMES 2009 test sample. The DAN predicts both text (printed in red under each text line) and layout entities (depicted as a graph on the right). Attention weights are colored given the last predicted layout token.}
    \label{fig:viz}
\end{figure*}

A visualization of the prediction for a test sample of the RIMES 2009 dataset is depicted in Figure \ref{fig:viz} \footnote{Full demo at  \url{https://www.youtube.com/watch?v=HrrUsQfW66E}}. On the left, attention weights of the last mutual attention layer are projected on the input image. The colors depend on the last predicted layout token. For visibility, the intensity of the colors is encoded for attention values between 0.02 and 0.25. The text prediction is added in red line by line, under the associated text line. The corresponding layout graph is depicted on the right, with each node corresponding to a text region in the input image. As one can note, even if the \modelacc{} is not trained using any segmentation label, it performs a kind of implicit segmentation in its process, which can be globally retrieved through the attention weights. 
As one can notice, the \modelacc{} performs a document recognition: it recognizes both text and layout.

In addition, the \modelacc{} is able to deal with slanted lines through its character-level attention mechanism. A prediction visualization for such example is depicted in Figure \ref{fig:viz_slanted_lines}. This time, we used one color per line for visibility. As one can note, the attention mechanism correctly follows the slope of the lines, leading to no error in prediction. This is an improvement compared to approaches based on line-level attention, which cannot handle such close and slanted lines.

\subsection{Ablation study}
\label{section:ablation}

\begin{table*}[ht]
    \caption{Ablation study of the DAN on the RIMES 2009 and READ 2016 datasets. Results are given in percentages, for the test sets, and for a 2-day training.}
    \centering
    \resizebox{\linewidth}{!}{
    \begin{tabular}{ l | c c c c | c c c c | c c c c}
    \hline
    & \multicolumn{4}{c|}{RIMES 2009 (single-page)}& \multicolumn{4}{c|}{READ 2016 (single-page)} & \multicolumn{4}{c}{READ 2016 (double-page)}\\
    & CER $\downarrow$ & WER $\downarrow$ & LOER $\downarrow$ & $\mathrm{mAP}_\mathrm{CER}$ $\uparrow$  & CER $\downarrow$ & WER $\downarrow$ & LOER $\downarrow$ & $\mathrm{mAP}_\mathrm{CER}$ $\uparrow$ & CER $\downarrow$ & WER $\downarrow$ & LOER $\downarrow$ & $\mathrm{mAP}_\mathrm{CER}$ $\uparrow$\\ 
    \hline
    \hline
    Baseline & 5.72 & 13.05 & \textbf{4.18} & \textbf{92.86} & \textbf{3.65} & 14.64 & 5.51 & 92.36 & 4.50 & 16.75 & 4.74 & \textbf{92.37}\\
    (1) No synthetic data  & 8.26 & 16.45 & 8.18 & 86.34 & 81.05 & 94.46 & 12.04 & 0.35 & 80.75 & 95.65 & 36.77 & 0.13\\
    (2) No curriculum for syn. data  & 7.59 & 16.48 & 6.63 & 88.92  & 4.28 & 15.41 & 5.62 & 91.66 & 78.89 & 92.05 & 15.42 & 0.00\\
    (3) No crop in curr. for syn. data & 5.84 & 13.73 & 4.42 & 91.94 & 100.00 & 100.00 & $>$ 100 & 0.00 & 100.00 & 100.00 & $>$ 100 & 0.00\\
    (4) No data augmentation  & 7.08 & 15.54 & 4.78 & 91.65  & 4.32 & 16.67 & 5.29 & 91.39 & 4.92 & 18.06 & 5.69 & 90.92\\
    (5) No curriculum dropout  & 5.83 & 14.41 & 4.36 & 92.09 & 3.92 & 14.85 & 5.51 & \textbf{93.13} & \textbf{4.23} & \textbf{16.12} & \textbf{3.68} & 92.26\\
    (6) No error in teacher forcing  & 8.09 & 15.12 & 5.91 & 89.24 & 7.51 & 21.87 & 4.95 & 83.51 & 85.78 & 99.51 & 42.35 & 10.73\\
    (7) No layout recognition & \textbf{5.30} & \textbf{12.46} & \xmark & \xmark  & 4.60 & 15.59 & \xmark & \xmark & 4.96 & 16.81 & \xmark & \xmark\\
    (8) No pre-training  & 71.42 & 87.48 & 18.46 & 12.72 & 4.47 & 16.32 & 4.72 & 90.52 & 5.84 & 20.47 & 5.81 & 88.24\\
    (9) No 1D positional encoding & 8.04 & 16.93 & 5.73 & 90.65 & 3.77 & \textbf{14.03} & 4.95 & 92.51 & 4.96 & 18.28 & 6.17 & 88.88\\
    (10) No 2D positional encoding & 12.43 & 20.83 & 8.42 & 89.81& 5.63 & 16.25 & \textbf{4.27} & 92.79 & 65.54 & 88.43 & 34.40 & 25.46\\
    \hline
    \end{tabular}
    }
    \label{table:ablation}
\end{table*}

We provide an extensive ablation study in Table \ref{table:ablation}. The evaluation is carried out on the test set of both RIMES 2009 and READ 2016 datasets, at single-page and double-page levels, after a 2-day training. We evaluated the proposed training approach through the independent removal of each of the training components we used. 

Experiments (1) to (3) are dedicated to the generation of synthetic documents at training time. In (1), we do not generate synthetic documents: the model is only trained on real documents. In (2), the synthetic documents do not follow a curriculum approach: the maximum number of lines per document is directly set to its upper bound ($l=l_\mathrm{max}$). In (3), the synthetic document images are not cropped below the lowest text line during the curriculum phase: they preserve the original image height. As one can note, the removal of either of these three aspects leads to poorer results for the RIMES 2009 dataset, for each metric. However, it prevents the model from adapting from training to evaluation for the READ 2016 dataset, except for (2) which only leads to degraded performance at single-page level. Indeed, when looking at the predictions, it turned out that the model learned the language from the training set: predictions are coherent (correct succession of words) but do not match the input document image. We assume that the complexity of the task, \textit{i.e.}, recognizing text and layout from whole documents directly, for (1) and (2), or from images of full size, for (3), combined with few training examples compared to RIMES 2009, lead to the learning of the training language as a shortcut to reduce the loss.
The removal of the data augmentation strategy during both pre-training and training, experimented in (4), leads to degraded results, for both datasets.

In (5), we do not use the curriculum dropout strategy either during pre-training or during training. This time, the results are rather mitigated, with an overall improvement at single-page level and a deterioration of the results at double-page level.
In experiment (6), we do not introduce any error in the teacher forcing strategy: we preserve the ground truth. The introduction of errors improves the results. We assume that training the model with errors helps it to deal with the errors made at prediction time. It also avoids learning the language from the training set in the case of the READ 2016 dataset at double-page level.

The layout tokens are removed from the ground truth in (7) leading to text recognition only. As one can notice, it leads to lower CER and WER for the RIMES 2009 dataset but higher ones for the READ 2016 dataset. This can be explained by the fact that the reading order is easier for RIMES 2009 than for READ 2016, especially at double-page level. We assume that recognizing the different layout entities enables to understand the spatial relationship between them and helps to learn the reading order. Annotations are always at the left of a body: after the prediction of an </annotation> token, the attention must be focused more on the right to predict a <body> token. 

In (8), the \modelacc{} is trained from scratch, without transfer learning from a prior pre-training step. Results are dramatically worse for the RIMES 2009 dataset. We assume that this is due to its irregular layout. Indeed, compared to READ 2016, RIMES 2009 shows a simpler reading order but with more variability in the layout. This way, we assume that for READ 2016, the reading order can be learned jointly with the feature extraction part, directly during the curriculum step. For RIMES 2009, the layout variability slows down the learning process of the feature extraction part, leading to slower convergence.

Finally, in (9) and (10), we evaluate the importance of the 1D and 2D positional encoding components. As one can notice, they both enable to improve the results, especially the 2D positional encoding. This is especially true for the READ 2016 dataset at double-page level. We assume this is due to the double-page nature of this dataset, which leads to a more complex layout, resulting in more important jumps from one character prediction to another. For example, from the last character of the left page to the first one of the right page.

\section{Limitations}
\label{section-limitations}
Although the \modelacc{} reaches promising results, this study faces some limitations.
The datasets used are rather specific: letters for RIMES and historical book pages for READ 2016, with homogeneous layouts for each one. The evaluation on the MAURDOR dataset, detailed in the Appendix, shows interesting results towards the recognition of more heterogeneous documents (mixing languages, printed and handwritten texts, as well as various backgrounds). However, the evaluation on MAURDOR does not take into account the layout recognition, due to the lack of annotations. It would be interesting to evaluate the whole approach on heterogeneous documents. However, for the time being, there is a lack of large-scale public datasets of documents including both layout annotation and text annotation. This is true for handwritten documents as well as for printed documents.

The proposed approach is limited to documents whose reading order is well-defined. Indeed, since the ground truth is a serialized representation of the document, it is required to know exactly in which order the layout entities must be read. This prevents the use of this approach for documents with multiple acceptable reading orders, such as many schemes or maps. This way, the training procedure, the ground truth annotations, and the evaluation protocol must be reconsidered to deal with these cases.

In its current implementation, the DAN outputs no geometric information about the location of the recognized symbols. Although some geometric information may be derived from the 2D attention maps, some oriented applications based on human interaction may require more accurate information for visualization purpose, for example. 

The obtained results are very dependent on the quality of the synthetic data, \textit{i.e.}, they must be close to the target dataset, notably in terms of layout. It means that the synthetic documents must cover the variety of document layouts in the target dataset. 
This could be a limitation for datasets containing very heterogeneous document layouts, since it would require to design a generic synthetic document generator.

The \modelacc{} is based on an autoregressive process. This is not a problem at training time since computations are parallelized through teacher forcing. However, this recurrence issue is significant at prediction time: it grows linearly with the number of tokens to be predicted. This can be an obstacle for an industrial application.

\section{Discussion}
\label{section-discussion}
As we have seen, the \modelname{} reaches great results on both text and layout recognition, whether it is at single-page or double-page level. Indeed, we showed that the \modelacc{} is robust to the dataset varieties: we used the same hyperparameters with two totally different datasets in terms of layout, language, color encoding as well as the number of training samples. 
We proposed an efficient training strategy, and we highlighted its impact on an extensive ablation study. This training strategy is based on synthetic line and document generation using digital fonts, to overcome the lack of training data. Pre-training is carried out on synthetic text lines only, avoiding using any segmentation annotations: this is a great advantage compared to state-of-the-art approaches. We showed that even with complex handwritings such as those of the READ 2016 dataset, for which the fonts' aspect is very far from the original writings, this approach remains effective.

The \modelacc{} learns the reading order through the transcription annotations. We observed an interesting effect linked to this. In the specific case of READ 2016 at double-page level, the page number is identical for both left and right pages. The \modelacc{} focuses on the same area to predict both numbers. Technically, this does not impact the performance, but it shows that the network has not fully learned the concept of reading. We assumed that this phenomenon would disappear by learning on samples with heterogeneous layouts.

We hope that this contribution will lead to the production of datasets at a lower cost: the \modelacc{} only needs the ordered transcription annotation and layout tags, without the need for any segmentation annotation.

\section{Conclusion}
\label{section-conclusion}
In this paper, we proposed a paradigm shift, from text recognition to document recognition. As opposed to the two-step paradigm used in the literature, this unified paradigm only relies on a single end-to-end architecture, reducing the adaptation effort from one dataset to another.

We proposed the \modelname{}, a new end-to-end segmentation-free architecture for the task of handwritten document recognition, following this unified paradigm. To our knowledge, it is the first end-to-end approach that can deal with whole documents, recognizing both text and layout. 

We showed that it is possible for the model to learn complex document-level reading order, without any physical segmentation annotation. Indeed, the model is trained only using text and layout tokens, reducing the annotation cost. By doing so, the document recognition task becomes very similar to a Natural Language Processing task: it does not deal with any physical information, but only with tagged text. This means that the physical and geometrical information are processed with a language supervision only, when training the \modelacc{}.

This new unified paradigm  is closer to the human way of reading a document. The text is sequentially recognized, character by character, through the character-level attention; lines are read one after the other, and the attention also jumps from one text block to the next until reaching the end of the document.

We introduced two new metrics to estimate the good recognition of text and layout altogether. We showed the efficiency of the \modelacc{} on two public datasets at single-page and double-page levels: RIMES 2009 and READ 2016, showing the robustness of the approach to different layouts. In addition, we showed that the proposed approach is robust enough to handle input images of various kinds: line, paragraph, and document.

Although the datasets used for evaluation are rather homogeneous, we assume that this approach could be generalized to heterogeneous documents without any problem by labeling a coherent reading order from one example to another.
The model is trained to output a structured sequence of characters and XML markups tokens. This way of serializing the document representation is modular enough to enrich the expected output with more information, such as named entities for instance, by adding dedicated tokens. This way, the proposed architecture seems promising to handle various tasks related to document understanding, whether they are handwritten or printed.

The main drawback of the approach is about the prediction time, which we aim at reducing in future works. 

\section*{Acknowledgment}
The present work was performed using computing resources of CRIANN (Regional HPC Center, Normandy, France) and HPC resources from GENCI-IDRIS (Grant 2020-AD011012155). This work was financially supported by the French Defense Innovation Agency and by the Normandy region.

\begin{figure}[H]
\centering
    \includegraphics[height=1.5cm]{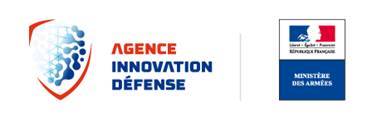}
    ~
    \includegraphics[height=1.5cm]{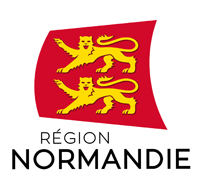}
    \label{data}
\end{figure}

\ifCLASSOPTIONcaptionsoff
  \newpage
\fi

\bibliographystyle{IEEEtran}
\bibliography{IEEEabrv,references.bib}

\begin{IEEEbiography}[{\includegraphics[width=1in,height=1.25in,clip,keepaspectratio]{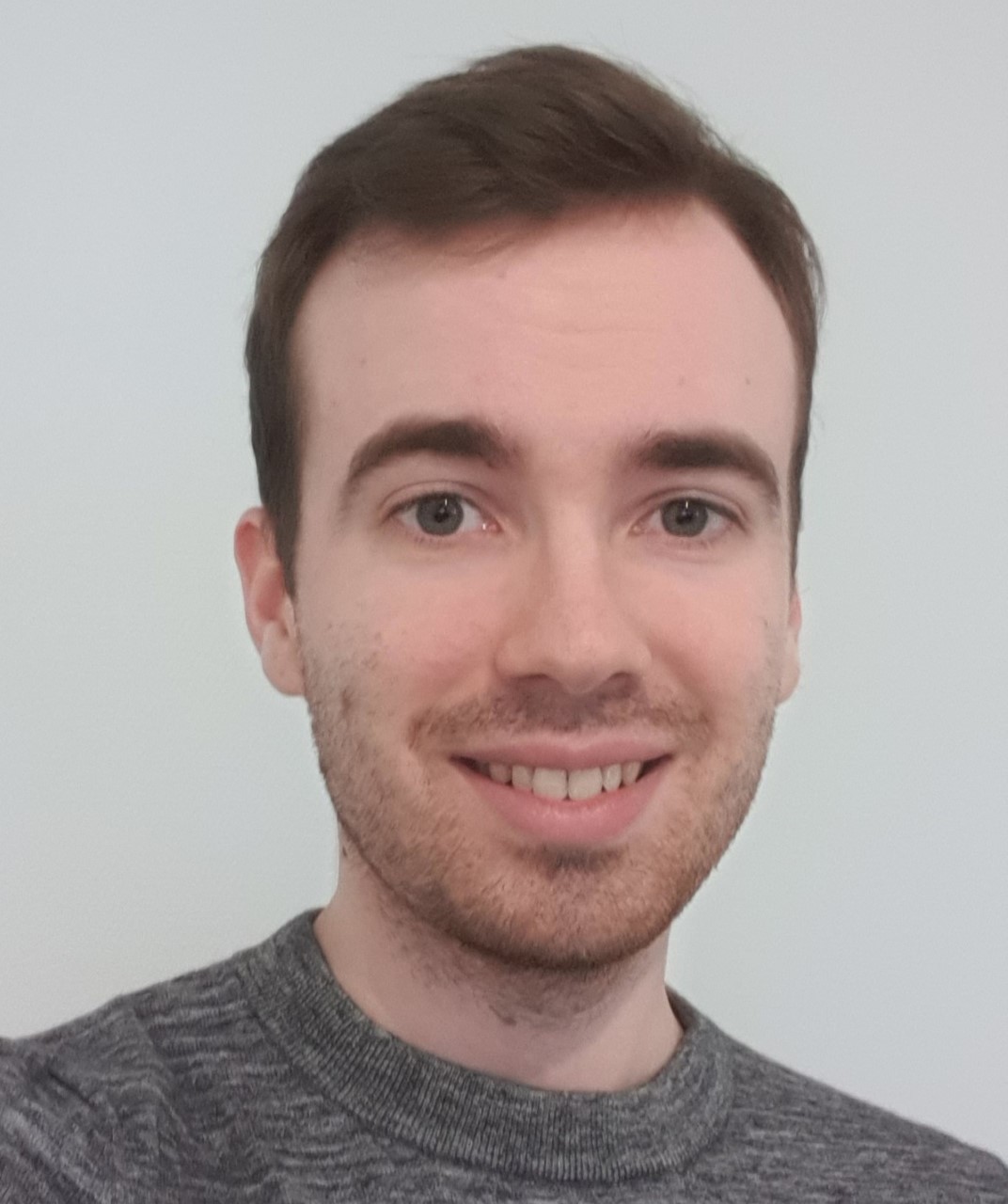}}]{Denis Coquenet}
received his Ph.D. in Computer Science in 2022 at the University of Rouen, France. His research interests include handwriting text recognition, document analysis and more globally computer vision and deep learning approaches.
\end{IEEEbiography}

\begin{IEEEbiography}[{\includegraphics[width=1in,height=1.25in,clip,keepaspectratio]{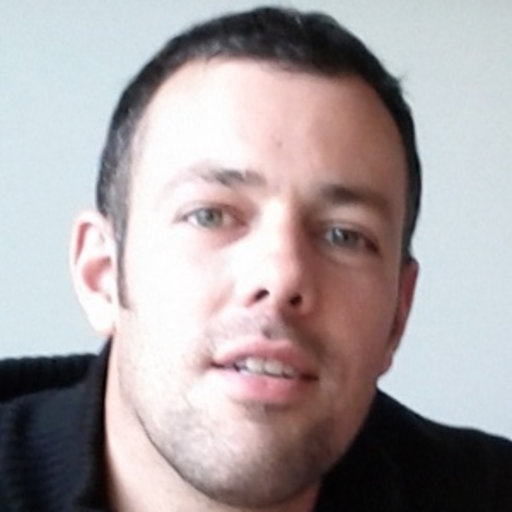}}]{Clément Chatelain}
is an Associate Professor in the Department of Information Systems Engineering at INSA Rouen Normandy, France. His research interests include machine learning applied to handwriting recognition, document image analysis and medical image analysis. His teaching interests include signal processing, deep learning and pattern recognition. In 2019, Dr. Chatelain received the French ability to conduct researches from the University of Rouen.
\end{IEEEbiography}

\begin{IEEEbiography}[{\includegraphics[width=1in,height=1.25in,clip,keepaspectratio]{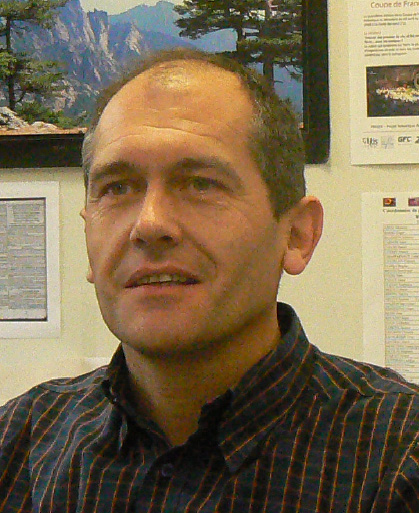}}]{Thierry Paquet}
was appointed Professor at the University of Rouen Normandie in 2002. He was the head of LITIS the research laboratory in Computer Science associated Rouen and Le Havre Universities, and Rouen INSA school of engineering. 
His research interests are machine learning, statistical pattern recognition, deep learning, for sequence modelling, with application to document image analysis and handwriting recognition. 
He has supervised 18 Ph.D. students on these topics and published more than 100 papers in international conferences and scientific journals. 
From 2008 to 2016, he was a member of the governing board of the French association for pattern recognition AFRIF. 
He was the president of the French association Research Group on Document Analysis and Written Communication (GRCE) from 2002 to 2010. 
\end{IEEEbiography}

\section*{Appendix A - Extended evaluation}

In order to have an idea of the robustness of the proposed approach on more complex data, we evaluate the \modelacc{} on the MAURDOR dataset \cite{MAURDOR}. 
The MAURDOR dataset consists of 10,000 annotated documents. We used the dataset from the second evaluation campaign, corresponding to 8,129 heterogeneous documents written in three languages (French, English and Arabic), and classified into 5 categories (C1: forms, C2: commercial documents, C3: private manuscripts correspondences, C4: private or professional correspondences, C5: others such as diagrams or drawings).

\begin{figure*}[h!]
\centering
    \begin{subfigure}{0.25\textwidth}
    \centering
    \includegraphics[width=\textwidth,frame]{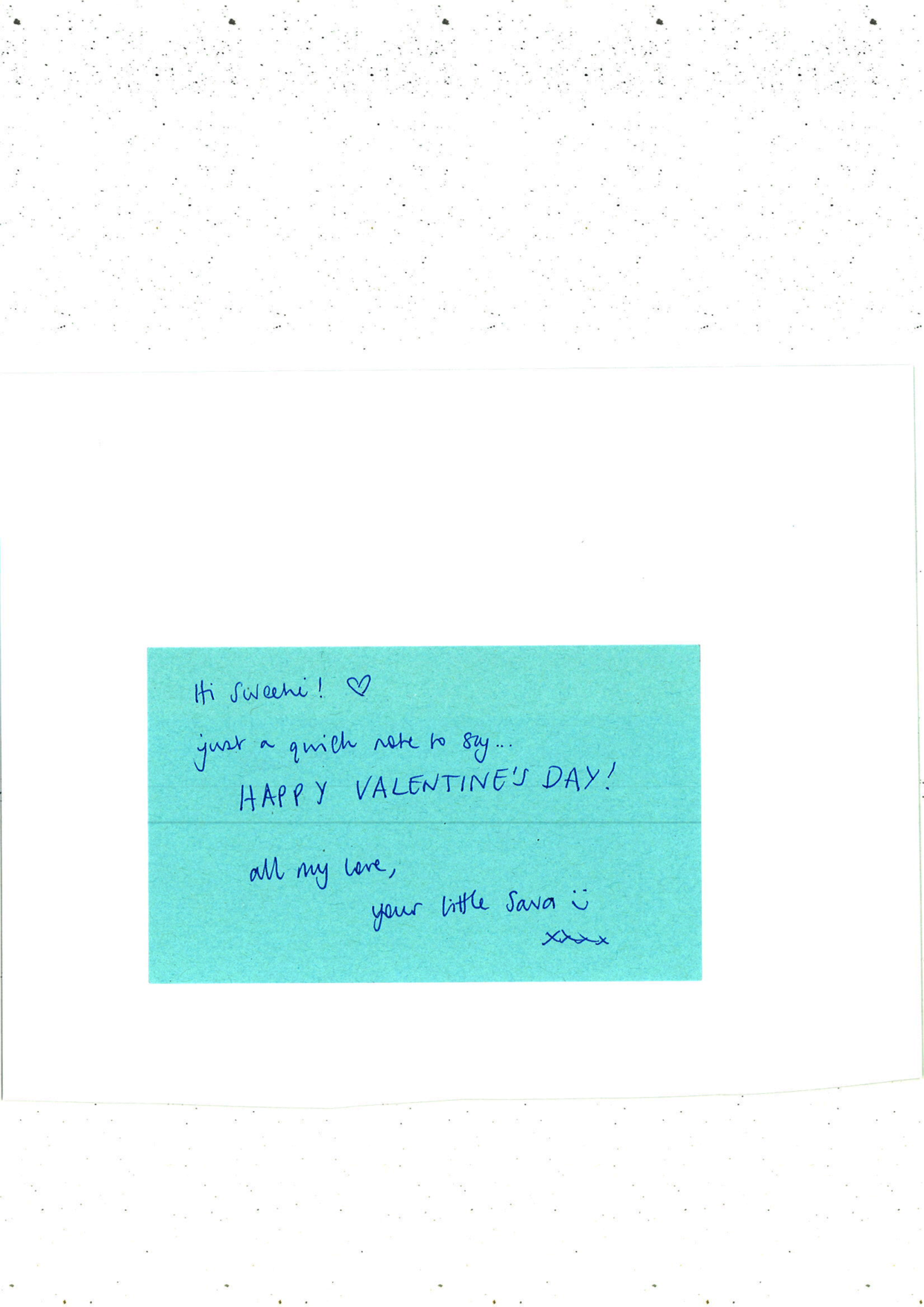}  
    \end{subfigure}
    ~
    \begin{subfigure}{0.25\textwidth}
    \centering
    \includegraphics[width=\textwidth,frame]{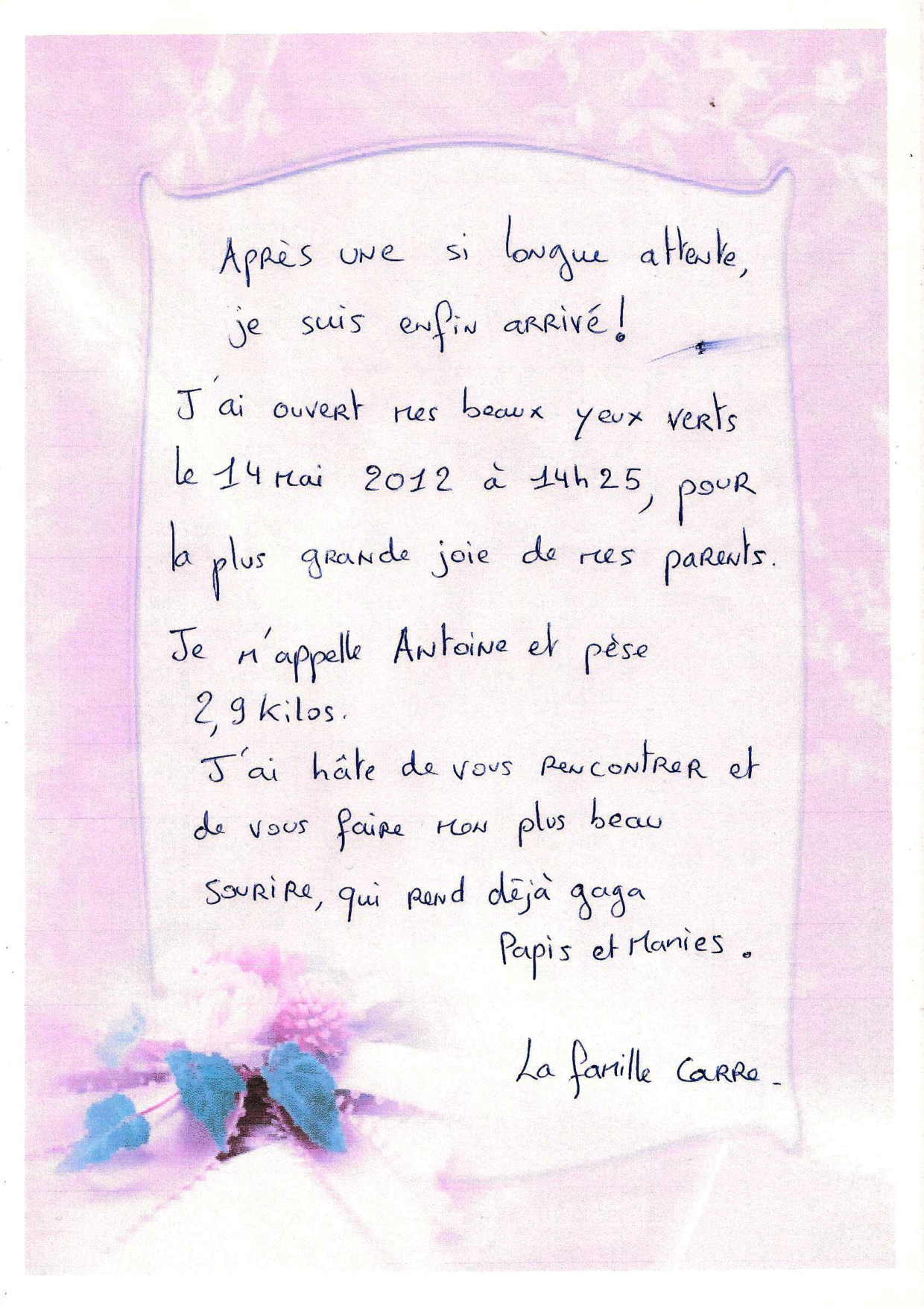} 
    \end{subfigure}
    ~
    \begin{subfigure}{0.25\textwidth}
    \centering
    \includegraphics[width=\textwidth,frame]{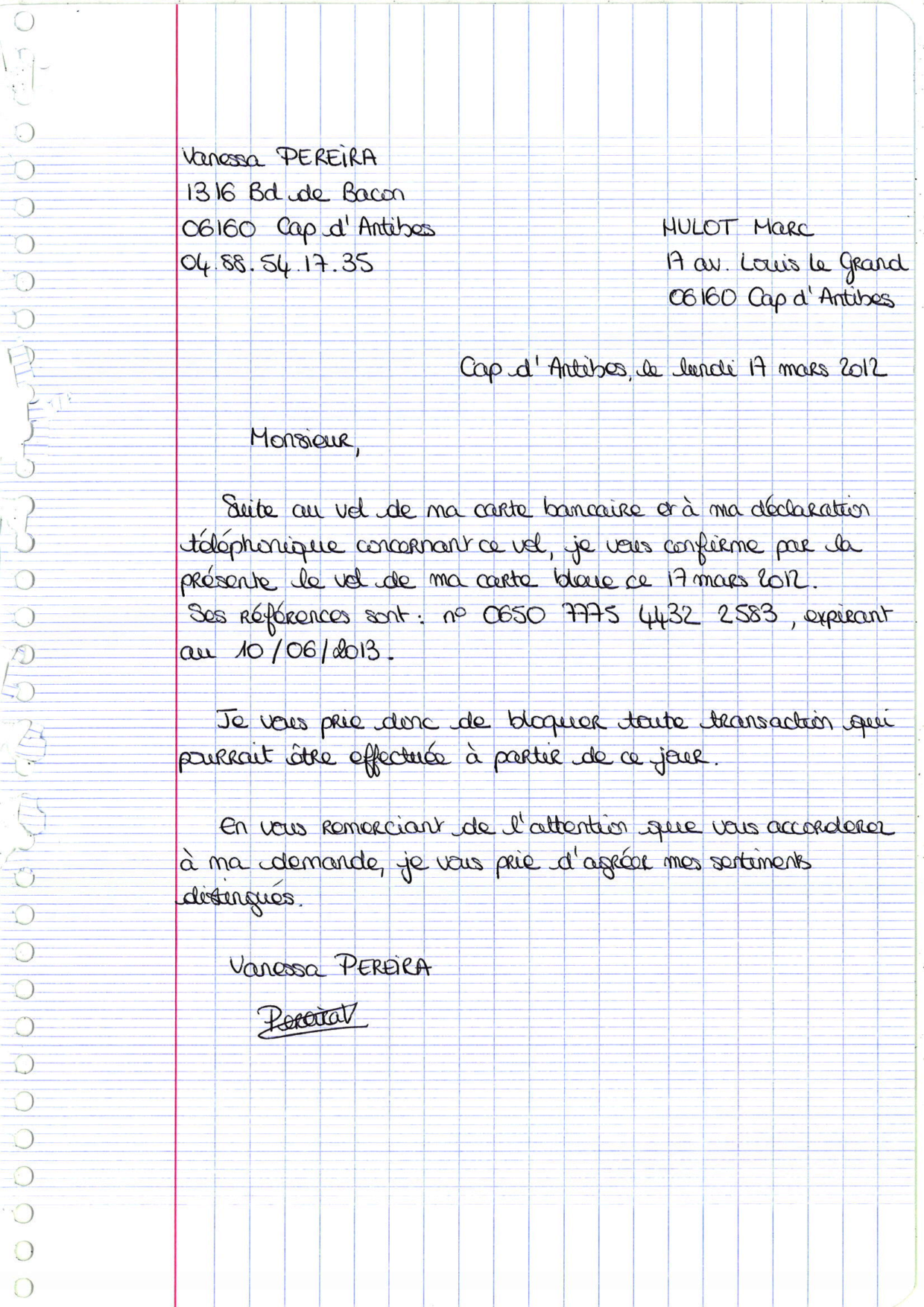}  
    \end{subfigure}
    \par\medskip
    \begin{subfigure}{0.25\textwidth}
    \centering
    \includegraphics[width=\textwidth,frame]{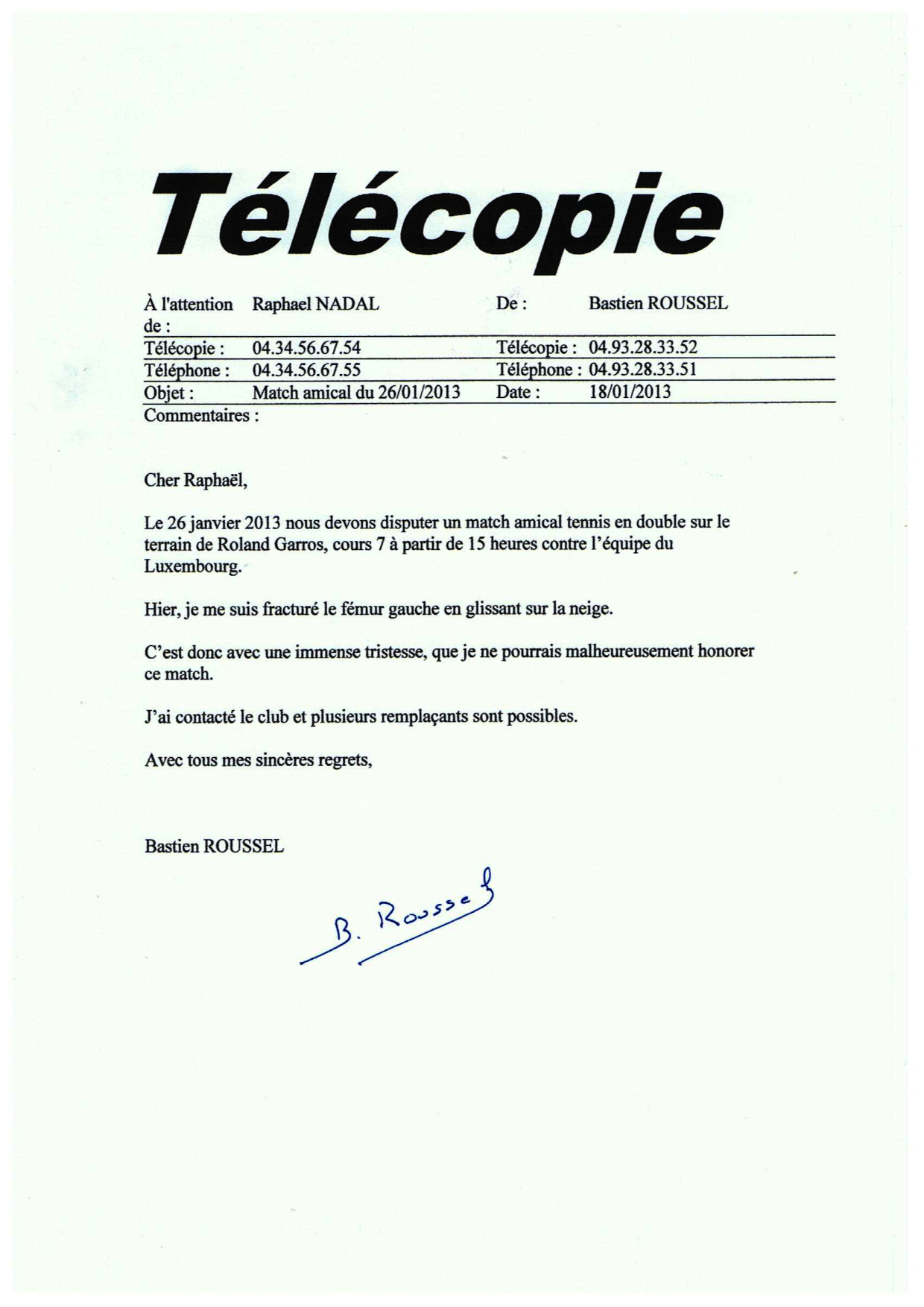}  
    \end{subfigure}
    ~
    \begin{subfigure}{0.25\textwidth}
    \centering
    \includegraphics[width=\textwidth,frame]{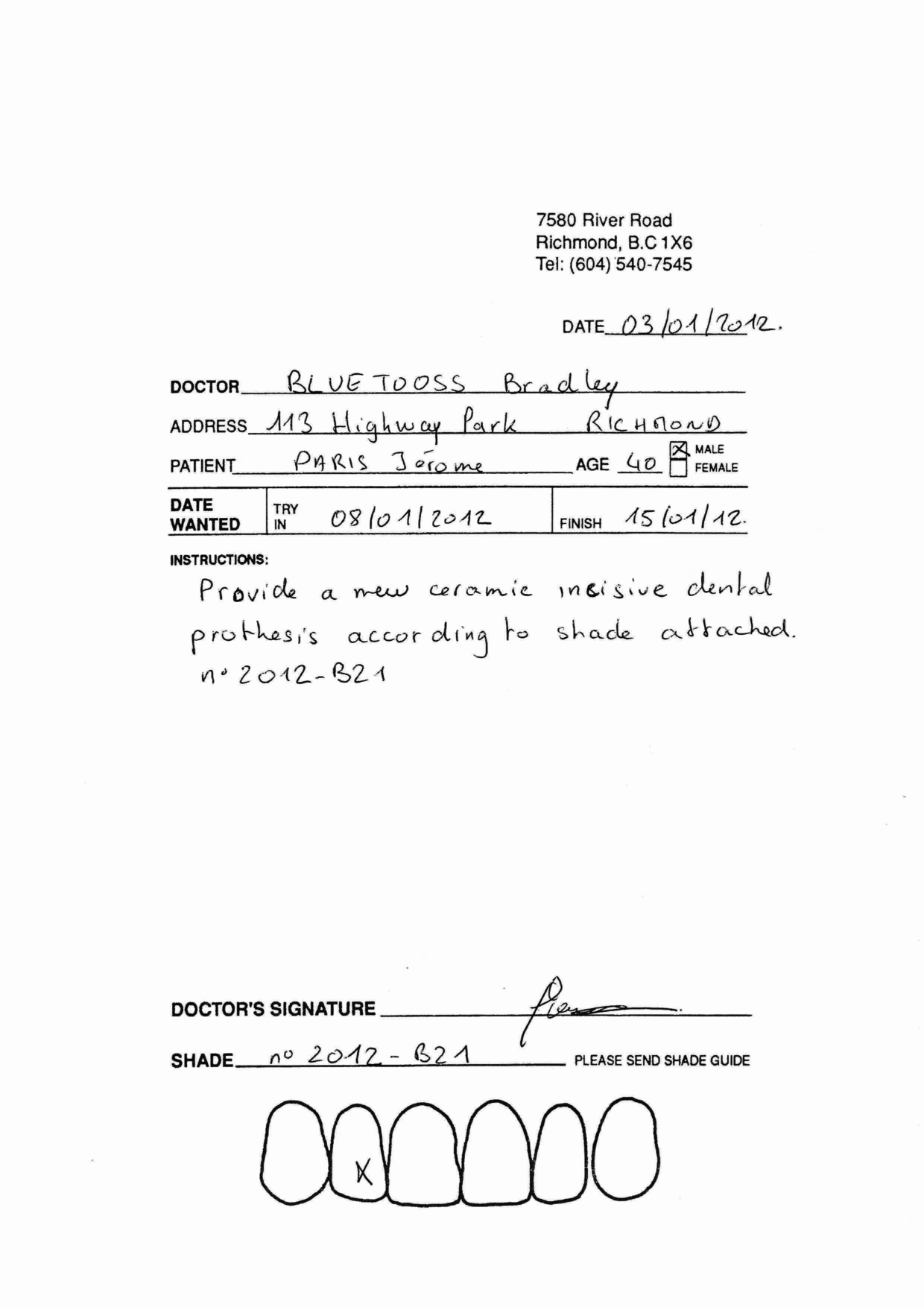}  
    \end{subfigure}
    ~
    \begin{subfigure}{0.25\textwidth}
    \centering
    \includegraphics[width=\textwidth,frame]{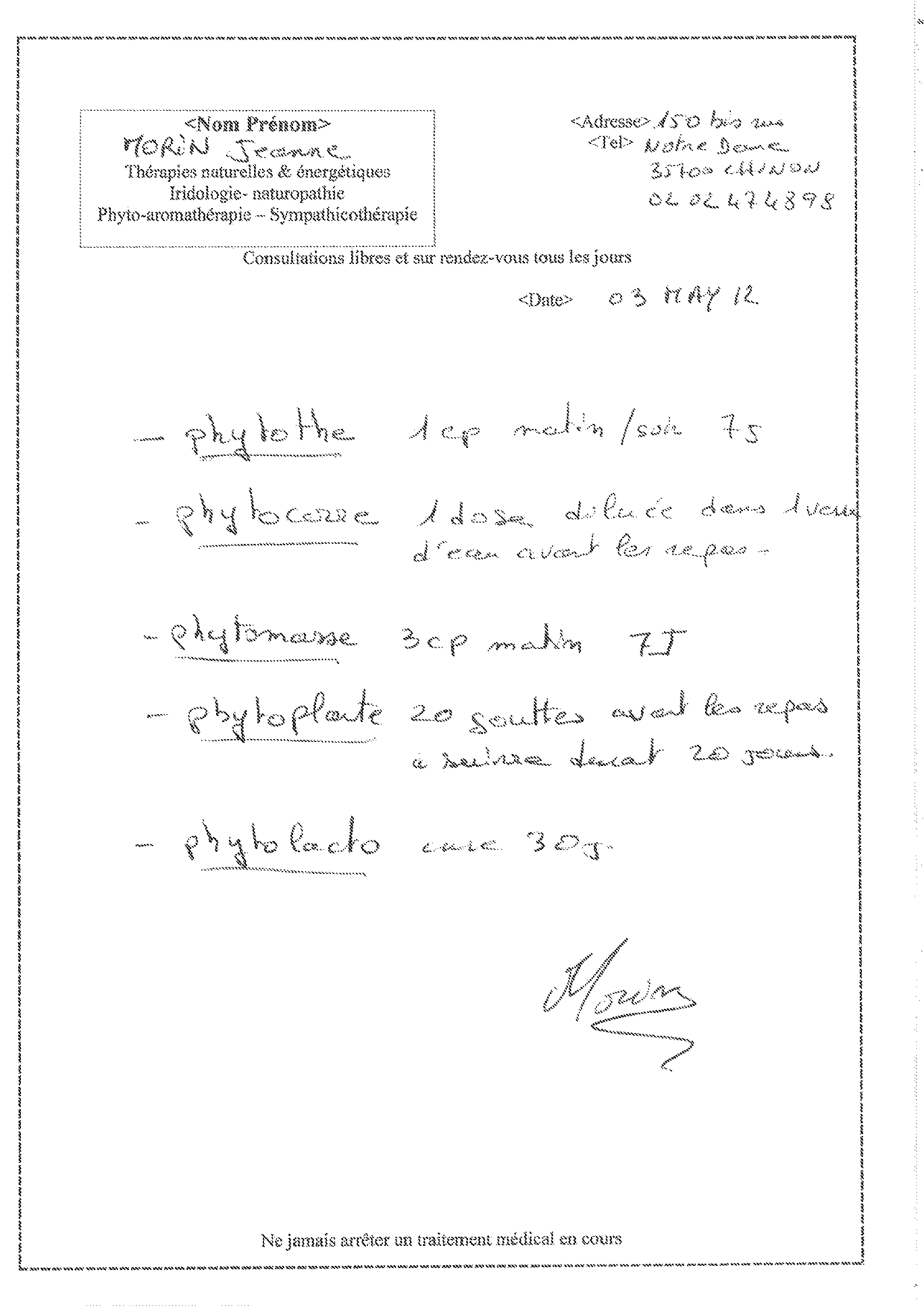}  
    \end{subfigure}

    \caption{Images from the MAURDOR dataset. Top: examples from C3. Bottom: examples from C4.}
\label{fig:dataset-maurdor}
\end{figure*}

Given that the reading order is not annotated homogeneously, we evaluate the \modelacc{} on documents from C3 and C4 only. Indeed, for these categories, the reading order can be generated automatically with fairly high confidence, based on the position of the different text regions. Examples from these two categories are shown in Figure \ref{fig:dataset-maurdor}. As one can note, the MAURDOR dataset is very challenging. Even in the same category, the documents look very different in terms of layout, background and non-textual items. In addition, writings can be typed, handwritten or a mix of both.

We used a sub-part of these two categories. We only used examples written in French and English to preserve the same rules regarding the reading order. Only the images with an A4 size are kept, in order to deduce the original resolution and adjust it to 150 DPI (for pre-processing). Images have been rotated when it was necessary to have the text in the right direction (the original images can be rotated by 90, 180 or 270 degrees).
This leads to the splits in training, validation and test described in Table \ref{table:split-maurdor}. The layout annotation of the MAURDOR dataset is not homogeneous, preventing us to use it for training. Indeed, in the case of a letter for instance, the text representing the information relative to the date and the location can be either considered as two different entities, merged together (associating two classes to the same text part), or the whole text can even be considered as only one of the two entities, ignoring the other one. This way, we only evaluate the text recognition task. We evaluate the \modelacc{} on the C3 and C4 categories, separately and jointly, to see the impact of grouping these two categories on the performance.

\begin{table}[h!]
    \caption{MAURDOR split in training, validation and test sets and associated number of tokens in their alphabet.}
    \centering
    \resizebox{\linewidth}{!}{
    \begin{tabular}{ l c c c c c}
    \hline
    \multirow{2}{*}{Dataset}  & \multirow{2}{*}{Training} & \multirow{2}{*}{Validation} & \multirow{2}{*}{Test} & \# char  & \# layout\\ 
     & & & & tokens & tokens\\
    \hline
    \hline     
    C3 & 1,006 & 148 & 166 & 134 & \xmark\\
    C4 & 721 & 111 & 114 & 127 & \xmark\\ 
    C3 \& C4 & 1,727 & 259 & 280 & 141 & \xmark \\
    \hline  

    \hline
    \end{tabular}
    }
    \label{table:split-maurdor}
\end{table}

We used exactly the same training strategy (pre-processing, pre-training, curriculum strategy based on synthetic data) as for the READ 2016 and RIMES 2009 datasets. To show the robustness of the proposed approach, we use exactly the same synthetic document generation process for both categories. Results on the test set of the MAURDOR dataset are given in Table \ref{table:maurdor}. As one can note, the obtained results are comparable for both categories: 8.62\% of CER for C3 and 8.02\% for C4. 
To our knowledge, the nearest work we can compare to is the system developed by A2iA \cite{Moysset2014}, which is the one giving the best results from the second evaluation campaign of the MAURDOR dataset. However, this work presents many differences compared to ours. The recognition task is evaluated from the isolated paragraphs (the paragraph segmentation step is not taken into account by the metrics). This way, the task is easier in that they do not have to handle the complexity of the reading order at document level; this enables them to use all the categories (C1 to C5) for training and evaluation. It has to be noted that the authors trained one specialized MDLSTM-based network per writing style and per language, leading to 6 networks, whereas we use a unique model. Contrary to our experiment, they also rely on a pre-training strategy using an external dataset, RIMES 2009, and on a 3-gram character language model to improve the results. The A2iA system reached a CER of 8.8\% for C3 and 6.2\% for C4. Considering all the differences between the two experiments, the \modelacc{} results seem interesting.

Combining both categories leads to an increase of the CER of about 3 points, which is significant. This could be explained by the variety of the input documents, making the reading order more difficult to deduce. As a matter of fact, since we cannot compute the $\mathrm{mAP}_\mathrm{CER}$, it is difficult to know how much is due to a misrecognition of the characters, and how much is due to a wrong reading order.

\begin{table}[ht]
    \caption{Evaluation of the DAN on the test set of the MAURDOR dataset.}
    \centering
    \resizebox{\linewidth}{!}{
        \begin{tabular}{ l c c c c c c}
        \hline
        Dataset &  CER $\downarrow$ & WER $\downarrow$ & Time & Input size & \# lines & \# num chars\\ 
        \hline
        \hline
        C3 & 8.62 & 18.94 & 5.76s & $3{,}316 \times 2{,}672$ & 16 & 481\\
        C4 & 8.02 & 14.57 & 7.66s & $3{,}508 \times 2{,}480$ & 22 & 706\\
        C3 \& C4 & 11.59 & 27.68 & 6.61s & $3{,}394 \times 2{,}594$ & 18 & 575\\
        \hline
        \end{tabular}
    }
    \label{table:maurdor}
\end{table}

We also specified the average inference time, input size, and the number of lines and characters per document for the test set. The inference time is given for a single GPU V100 (32Go). As one can note, this is consistent with the inference times of RIMES 2009 and READ 2016.

Table \ref{table:maurdor-details} details the performance for each writing style (printed, handwritten or a mix of both) and for each language (French, English or a mix of both). The number of corresponding samples is also indicated to give a confidence level on the metrics. As one can note, the results are better for printed than for handwritten documents, which was expected. One should note that the training set is rather well-balanced with respect to the test set in terms of writing style: 2 printed, 522 handwritten and 482 mix for C3, and 345 printed, 7 handwritten and 369 mix for C4. 

\begin{table*}[ht]
    \caption{Evaluation of the DAN on the test set of the MAURDOR dataset per language and writing type.}
    \centering
    \resizebox{\linewidth}{!}{
        \begin{tabular}{ l c | c c c c | c c c c | c c c c | c c c c  }
        \hline
        Dataset & Metric & \multicolumn{4}{c|}{Printed} & \multicolumn{4}{c|}{Hanwdritten} & \multicolumn{4}{c|}{Mix}& \multicolumn{4}{c}{All}\\ 
        & & FR & EN & Mix & All & FR & EN & Mix & All & FR & EN & Mix & All & FR & EN & Mix & All\\
        \hline
        \hline
        \multirow{3}{*}{C3} & \# samples & 0 & 0 & 0 & 0 & 42 & 55 & 0 & 97 & 63 & 4 & 2 & 69 & 105 & 59 & 2 & 166\\
        & CER (\%) & \xmark & \xmark & \xmark & \xmark & 6.13 & 13.39 & \xmark & 8.57 & 7.86 & 8.46 & 10.46 & 7.98 & 7.17 & 12.99 & 10.46 & 8.62\\
        & WER (\%) & \xmark & \xmark & \xmark & \xmark & 14.83 & 30.69 & \xmark & 20.50 & 17.10 & 20.23 & 25.96 & 17.50 & 16.22 & 29.89 & 25.96 & 18.94\\
        \hline
        \multirow{3}{*}{C4} & \# samples & 47 & 9 & 2 & 58 & 0 & 1 & 0 & 1 & 35 & 18 & 2 & 55 & 82 & 28 & 4 & 114\\
        & CER (\%) & 5.39 & 0.86 & 10.93 & 5.05 & \xmark & 12.94 & \xmark & 12.94 & 10.67 & 12.89 & 12.79 & 11.26 & 7.42 & 9.17 & 12.01 & 8.02\\
        & WER (\%) & 9.94 & 2.12 & 12.64 & 9.05 & \xmark & 35.04 & \xmark & 35.04 & 18.36 & 24.61 & 23.61 & 20.45 & 13.42 & 17.60 & 18.98 & 14.57\\
        \hline
        \multirow{3}{*}{C3 \& C4} & \# samples & 47 & 9 & 2 & 58 & 42 & 56 & 0 & 98 & 98 & 22 & 4 & 124 & 187 & 87 & 6 & 280\\
        & CER (\%) & 8.49 & 0.26 & 59.83 & 9.55 & 6.87 & 36.01 & \xmark & 16.96 & 9.20 & 12.59 & 13.11 & 9.90 & 8.51 & 21.14 & 27.05 & 11.59\\
        & WER (\%) & 13.96 & 2.95 & 58.71 & 14.44 & 17.84 & 124.51 & \xmark & 56.87 & 18.42 & 22.26 & 24.09 & 19.23 & 17.10 & 68.27 & 34.52 & 27.68\\
        \hline
        \end{tabular}
    }
    \label{table:maurdor-details}
\end{table*}

The results are better for the French language than for the English language. This can be explained by the training set which includes more French samples: 698 French, 287 English and 21 mix for C3, and 479 French, 232 English and 10 mix for C4. 

These results are of the same order of magnitude as those of the A2iA system which reached 6.8\% of CER for French/printed, 8.1\% for English/printed, 9.4\% for French/handwritten and 17.5 for English/handwritten. It has to be noted that, since the A2iA system takes paragraph images as input, there cannot be a mix of handwritten and printed text among the same example in their case. This way, a document mixing printed and handwritten text in our experiment is considered as handwritten paragraphs and printed paragraphs for them; this should be taken into account when comparing the metrics. It is the same for the language mixes.

When mixing both categories (C3 and C4), one can note that the results are globally worse. Indeed, for a few examples, the model gets confused with the reading order, leading to a high CER. We noticed a single example for which the model does not stop the recurrent process by predicting the end-of-prediction token, but by waiting for the fixed limitation of 3,000 predicted tokens. This dramatically increases the global CER and WER values, leading to a WER greater than 100\% for the English handwritten documents. 

These experiments, although limited by the lack of layout annotations, show promising results for the end-to-end recognition of complex, heterogeneous documents.


\section*{Appendix B - List of acronyms}

Table \ref{tab:acronyms} provides the list of the acronyms used in this paper.

\begin{table}[h!]
    \caption{List of acronyms.}
    \centering
    \begin{tabular}{r|l}
     AP & Average Precision\\
     AUC & Area Under the Curve \\
     BERT & Bidirectional Encoder Representations from Transformers\\
     CE & Cross-Entropy\\
     CER & Character Error Rate \\
     CNN & Convolutional Neural Network\\
     CTC & Connectionist Temporal Classification\\
     DAN & Document Attention Network\\
     DLA & Document Layout Analysis\\
     FCN & Fully Convolutional Network\\
     GED & Graph Edit Distance\\
     GPU & Graphics Processing Unit \\
     HDR & Handwritten Document Recognition\\
     HTR & Handwritten Text Recognition\\
     IoU & Intersection over Union\\
     KIE & Key Information Extraction\\
     LOER & Layout Ordering Error Rate \\
     LSTM & Long Short-Term Memory\\
     mAP & mean Average Precision\\
     MD-LSTM & Multi-Dimensional Long Short-Term Memory\\
     OCR & Optical Character Recognition\\
     PPER & Post-Processing Edition Rate \\
     VDU & Visually-rich Document Understanding\\
     VQA & Visual Question-Answering\\
     WER & Word Error Rate \\
     XML & eXtensible Markup Language\\
     
    \end{tabular}
    \label{tab:acronyms}
\end{table}

\section*{Appendix C - List of symbols}
Table \ref{tab:symbols} provides the list of the symbols used in this paper.

\begin{table*}[h!]
    \caption{List of symbols.}
    \centering
    \begin{tabular}{c|l}
        \multicolumn{2}{l}{\textbf{Architecture and training:}}\\
        \multicolumn{2}{l}{}\\
         $\mb{X}$ & Input image  \\
         $\mseq{y}$ &  Expected output sequence\\
         
         $\mset{A}$ & Set of character tokens \\
         $\mset{S}$ & Set of layout tokens \\
         $\mset{D}$ & Set of predictable tokens\\
         
         $\mathrm{<sot>}, \mathrm{<eot>}$ & Special start-of-transcription/end-of-transcription token\\
         
         $\mb{f}_\mathrm{2D}$ & 2D features extracted from the encoder\\
         $\mb{f}_\mathrm{1D}$ & Flattened features with positional encoding\\
         $\mb{o}_t$ & Output of the transformer decoder at iteration $t$ (character-level representation)\\
         $\mb{p}_t$ & Probabilities for each predictable token, at iteration $t$\\
         $\hat{\mseq{y}}_t$ & Predicted token at iteration $t$\\
         $\mb{q}_t$ & Query at iteration $t$\\ 
         
         $\mathrm{PE}_\mathrm{2D}$ & 2D positional encoding tensor\\
         $\mathrm{PE}_\mathrm{1D}$ & 1D positional encoding tensor\\
         $w_k$ & Pulsation of sine and cosine functions, for the channel $k$\\
         
         $\mb{e}$ & Embedding tensor\\
         
         $H, W, C$ & Dimensions (height, width, number of channels) of $\mb{X}$\\
         $H_f, W_f, C_f$ & Dimensions (height, width, number of channels) of $\mb{f}_\mathrm{2D}$\\
         $L_y$ & Length of $\mseq{y}$\\
         $L_\mathrm{max}$ & Maximum number of iterations for the decoding process\\
         $d_\mathrm{model}$ & Dimension of a token embedding\\
         
         $\mset{L}$ & Loss used to train the model\\
         $\mset{L}_\mathrm{CE}$ & Cross-entropy loss\\
        
        \multicolumn{2}{l}{}\\
        \multicolumn{2}{l}{\textbf{Curriculum learning:}}\\
        \multicolumn{2}{l}{}\\
        
        $\tau_t$ & Dropout rate after $t$ weight updates\\
        $\bar{\tau}$ & Dropout rate after the curriculum dropout phase \\
        $T$ & Estimated number of weight updates during training \\
        $\mathcal{D}_\mathrm{doc}$ & Dataset at document level \\
        $\mathcal{D}_\mathrm{line}$ & Dataset at line level \\
        $\mathcal{F}$ & Set of fonts\\
        $\mathcal{Y}$ & Set of ground truth transcriptions \\
        $\mathcal{C}$ & Set of layout classes\\
        $c$ & Layout class \\
        $f$ & Font\\
        $s$ & Stylesheet used to generate synthetic documents\\
        
        $\mb{D}$ & Synthetic document image\\
        $\mb{i}_k, \mb{i}_\mathrm{entity}$ & Image of the current text line/layout entity\\
        $l_\mathrm{max}$ & Maximum number of lines per page in synthetic documents\\
        $l$ & Current maximum number of lines per page in synthetic documents\\
        $l_\mathrm{doc}$ & Expected number of lines per page in the current synthetic document\\
        $l_\mathrm{current}$ & Current number of lines per page in the current synthetic document\\
        $l_\mathrm{entity}$ & Number of lines in the current layout entity\\

        \multicolumn{2}{l}{}\\
        \multicolumn{2}{l}{\textbf{Metrics:}}\\
        \multicolumn{2}{l}{}\\
         $K$ & Number of examples in the dataset\\
         $\hat{\mseq{y}}^\mathrm{text}, \mseq{y}^\mathrm{text}$ & Text tokens of the prediction/ground truth sequence\\
         $\hat{\mseq{y}}^\mathrm{layout}, \mseq{y}^\mathrm{layout}$ & Layout tokens of the prediction/ground truth sequence\\
         $\hat{\mseq{y}}^\mathrm{graph}, \mseq{y}^\mathrm{graph}$ & Graph representation of the predicted/ground truth layout tokens\\
         $\mb{y}^{\mathrm{text}}_{\mathrm{len}}$, $\mb{y}^{\mathrm{layout}}_{\mathrm{len}}$ & Number of text/layout tokens in the ground truth\\
         $\mathrm{d}_\mathrm{lev}$ & Levenshtein distance \\
         $\mathrm{GED}$ & Graph edit distance \\
         $n_{\mathrm{e}}, n_{\mathrm{n}}$ & Number of edges/nodes in the graph\\
         $n_{\mathrm{ppe}}$ & Number of post-processing edition operations\\
         $r, p, p_\mathrm{interp}$ & Recall, precision and interpolated precision \\
         $\theta_\mathrm{min}$, $\theta_\mathrm{max}$, $\Delta\theta$ & Minimum, maximum, and step for the CER threshold of $\mathrm{AP}_{\mathrm{CER}}$ computation\\
         $\mathrm{len}_c$ & Number of characters corresponding to the class $c$ in the ground truth\\
         $\mathrm{AP}_{\mathrm{CER}_c}^{5:50:5}$ & Average precision averaged for CER thresholds from 5\% to 50\% with a step of 5\%\\

    \end{tabular}
    
    \label{tab:symbols}
\end{table*}

\end{document}